\documentclass[letterpaper]{article} % DO NOT CHANGE THIS
\usepackage[preprint]{aaai2027} % DO NOT CHANGE THIS
\usepackage[hyphens]{url}  % DO NOT CHANGE THIS
\usepackage{graphicx} % DO NOT CHANGE THIS
\usepackage{natbib}  % DO NOT CHANGE THIS AND DO NOT ADD ANY OPTIONS TO IT
\usepackage{caption} % DO NOT CHANGE THIS AND DO NOT ADD ANY OPTIONS TO IT
\usepackage{algorithm}
\usepackage{algorithmic}
\usepackage{float}
\usepackage{newfloat}
\usepackage{listings}
\DeclareCaptionStyle{ruled}{labelfont=normalfont,labelsep=colon,strut=off} % DO NOT CHANGE THIS
\floatstyle{ruled}
\newfloat{listing}{tb}{lst}{}
\floatname{listing}{Listing}

\usepackage{booktabs}
\usepackage{amssymb}            % Defines common symbols like \mathbb R
\usepackage{mathtools}          % Extends amsmath, providing common math tools
\usepackage{multirow}
\usepackage{subcaption}         % Allows for the use of subfigures and subcaptions
\usepackage[table]{xcolor}  % provides \cellcolor, \rowcolor
\usepackage{array}

\definecolor{bestblue}{RGB}{190,225,255}
\newcommand{\best}[1]{\cellcolor{bestblue}#1}
\newcommand{\second}[1]{\cellcolor{gray!15}#1}
\title{Steerable Pluralistic Policy Alignment from Anonymized Crowd Feedback}
\author{
    Akansha Kalra\textsuperscript{\rm 1}\corresponding,
    Ellen R. Novoseller\textsuperscript{\rm 2},
    Vernon J. Lawhern\textsuperscript{\rm 2},
    Daniel S. Brown\textsuperscript{\rm 1}
}

\affiliations{
    \textsuperscript{\rm 1}Kahlert School of Computing, University of Utah, Salt Lake City, USA\\
    \textsuperscript{\rm 2}U.S. Army DEVCOM Army Research Laboratory, Aberdeen Proving Ground, MD, USA\\
    {akansha.kalra@utah.edu, \{ellen.r.novoseller.civ, vernon.j.lawhern.civ\}@army.mil, daniel.s.brown@utah.edu}
}

\begin{document}

\maketitle

% \begin{abstract}
% Learning from diverse human feedback and personalizing AI models to each user is a hard challenge. 
% Most research on reinforcement learning from human feedback (RLHF) either assumes all preference annotators share the same underlying reward distribution and treat divergent preferences as noise or assume prior knowledge about annotator identity to learn diverse behaviors conditioned on the specific annotator ID.
% We develop a framework that delivers personalized pluralistic alignment while learning from  diverse, potentially conflicting preferences without requiring prior knowledge about annotator identity. 
% Our framework learns a shared basis of preference directions from heterogeneous crowd feedback by combining trajectory representation learning with a PCA decomposition of preference signals.
% These basis eigenvectors paired with adaptation queries define the personalized reward which is then used to optimize a downstream policy. 
% By leveraging heterogeneous preferences, our approach produces personal alignment that generalizes beyond the offline dataset to out-of-distribution new users and attains competitive RL and test performance with reduced computational costs, and is flexible to varying  adaptive query budgets.

% \end{abstract}

\begin{abstract}
% Learning from diverse human feedback and personalizing AI
% models to each user is a hard challenge. Most research on
% reinforcement learning from human feedback (RLHF) 
% assumes all preference annotators share the same underlying
% reward distribution and treat divergent preferences as noise.
Pluralistic alignment requires AI systems to learn from diverse populations whose preferences may reflect heterogeneous or even conflicting objectives, and to personalize their behavior to individual users. 
Existing approaches in pluralistic alignment often assume access to annotator identities and rely on powerful pretrained representations to model individual preference variations.
% The latter assumption is inherently natural for language models, which typically undergo large-scale pretraining and supervised fine-tuning before preference-based alignment. 
% Embodied control presents a significantly harder challenge as the underlying task objectives as well as the user preference distribution must be learned jointly
% from the crowd preference dataset using only the self-supervised representation. 
% without comparably pretrained task-specific representations, an agent must learn meaningful trajectory features while simultaneously recovering structure from heterogeneous and potentially conflicting human preferences. 
We propose Preference-PCA Pluralistic Learning (P3L), a novel framework for pluralistic alignment that jointly learns trajectory embeddings and the shared basis of preference directions via PCA decomposition from heterogeneous crowd feedback without requiring annotator identities.
% Given pairwise preference crowd dataset, P3L constructs preference-difference embeddings and uses PCA to obtain preference eigenvectors.
% Our framework uses learns a shared basis of preference directions from heterogeneous crowd feedback by combining self-supervised
% trajectory representation learning with a PCA decomposition
% of preference queries.
At test time, our approach enables rapid and accurate personalized reward learning and downstream policy optimization.
% These basis eigenvectors, paired with adaptation queries at test time,  define the personalized reward which is then used to optimize a downstream policy.  
% For a new user, a small number of adaptive preference queries determines weights over this basis, yielding a personalized reward function that can directly optimize a downstream policy. 
% Across multi-objective embodied-control environments, this representation supports strong personalization to both observed and previously unseen users, competitive RL performance, flexible adaptation budgets, and substantially reduced training cost.
Experiments across multiple embodied control tasks demonstrate that P3L enables steerable personal alignment that generalizes beyond the offline dataset to out-of-distribution new users and attains competitive RL and test performance, while substantially reducing computational cost and supporting variable adaptation query budgets.

% Across mulitple embodied control tasks, P3L achieves personalized alignment that generalizes beyond the offline training population to out-of-distribution users, while maintaining competitive downstream RL performance, reducing training cost, and supporting flexible test-time adaptation.

\end{abstract}

\section{Introduction}
As we witness growing adoption of AI systems and robotic models for downstream applications such as in industrial automation~\cite{tang2023industreal,kernbach2026behavioral}, assistive healthcare~\cite{bouzida2024carmen,lamsey2024stretch}, and household robotics~\cite{shafiullah2023bringing}, being able to steer and align these models towards an end-user's individual preference is seen as a major challenge~\citep{ji2023ai,casper2023open}. 
One of the most popular approaches for aligning such models is Reinforcement Learning from Human Feedback (RLHF).  

RLHF learns a reward function from human preferences and then optimizes an RL policy under the learned reward function ~\citep{christiano2017deep,ouyang2022training,kaufmann2024survey}. 
While this approach has been highly successful, it has been shown to be susceptible to over optimization~\citep{anwar2024foundational} and reward misidentification~\citep{tien2023causal}, and can become easily misaligned across diverse users in a crowd ~\citep{dpl_siththaranjan2024distributional,vpl_poddar2024personalizing,chhan2024crowd}. 
Recent work has tried to solve this personalization challenge by either learning multiple reward functions from divergent crowd-sourced preferences ~\citep{dorsa-myers22a,Pref_pulkit} or by assuming knowledge about the annotators who provide the preference labels, such as the annotator's id ~\citep{vpl_poddar2024personalizing} . 
However, this assumption limits applicability in many real-world settings where annotator identities are unavailable, anonymized or cannot be reliably associated with preference labels. 
Moreover, VPL has been shown to suffer from posterior collapse~\citep{kim2026swapguided_iclr}, a failure mode that we also observe empirically.
% (For more details, please refer to Appendix~\ref{app:vpl_posterior_collapse}. 
This raises a central challenge: \textit{Can we learn from heterogeneous crowd preferences—potentially reflecting conflicting objectives—without annotator identity, while still enabling efficient test-time personalization to new users in embodied control tasks?}

Our key idea is to decompose heterogeneous crowd preferences into a small set of shared preference directions. 
This reframes pluralistic alignment as estimating user-specific weights over a shared preference basis representation rather than learning separate reward models for each user.
This representation enables personalized reward functions—and corresponding policies—to be constructed for new users at test time without requiring annotator identities, allowing rapid policy personalization from a small number of adaptive queries.
Building on the preference vector framework~\citep{pca}, which first proposed to represent human preferences as directions in a latent representation space in the context of LLMs, we extend and test this idea in embodied control scenarios. Our approach jointly learns trajectory representations and preference directions from heterogeneous crowd feedback. Specifically, we train an end-to-end unsupervised representation encoder that embeds trajectories and we use these embeddings to extract a basis of preference eigenvectors capturing variation in user objectives.

Unlike prior work such as Variational Preference Learning (VPL)~\citep{vpl_poddar2024personalizing}, which assumes access to annotator identities to infer user-specific rewards, our framework does not require annotator information. However, our framework can incorporate annotator identities when available, enabling it to operate under both anonymized and privileged settings.
Overall, our approach enables a simple, flexible, and effective framework for pluralistic alignment in embodied control.

We summarize our contributions as follows:
% \begin{itemize}
    % \item
 \textbf{(1) Learning preference bases from heterogeneous feedback.} 
    We introduce a framework that learns a shared basis of preference directions from heterogeneous, potentially conflicting crowd preferences. Our method leverages an unsupervised trajectory encoder to construct latent trajectory embeddings and extracts preference eigenvectors that capture variation in user objectives without requiring annotator identities.
    % \item 
\textbf{(2) Personalized policy alignment.} 
    We demonstrate that the learned preference basis enables test-time construction of personalized reward functions and policies, allowing alignment to individual users across multiple objectives.
    % \item
  \textbf{(3) Empirical evaluation.} 
    Across multi-objective control environments, our method achieves strong alignment to both in-distribution and out-of-distribution users. Compared to prior work such as VPL~\citep{vpl_poddar2024personalizing}, it achieves higher test-time preference accuracy and competitive RL performance while significantly reducing training cost and allowing flexible adaptation to varying numbers of test-time queries.

\section{Related Work}
\paragraph{Reinforcement Learning from Human Feedback (RLHF)} addresses the challenge of specifying reward functions by instead learning them from human feedback. Human feedback can take several forms, including pairwise preferences~\citep{christiano2017deep,vpl_poddar2024personalizing}, rankings~\citep{brown2019extrapolating,dorsa-myers22a}, and corrective feedback~\citep{bajcsy17a,mehta2024unified}. 
% Among these approaches, learning reward functions from pairwise preferences has become one of the most widely adopted paradigms.
% However, prior work has shown that policies optimized under learned reward functions can exhibit unintended or undesirable behaviors. Early studies demonstrated such effects in Atari environments when optimizing rewards learned from human preferences~\citep{christiano2017deep,brown2020safe,ibarz2018reward}. More recent work has further shown that high accuracy on held-out preference queries does not necessarily translate to correct downstream RL behavior due to spurious correlations or reward misidentification in the learned objectives~\citep{tien2023causal,kalra2024can}. 
Most RLHF methods learn reward functions from pairwise preferences and assume that this preference data originates from a single underlying reward function, effectively treating disagreement among annotators as noise. Some work attempts to account for imperfect human feedback. For example, \citet{bpref2021} introduced stochastic preference models to capture noisy or inconsistent human judgments. Nevertheless, these approaches still assume that annotators ultimately share a common objective.
By contrast, real-world preference data is often collected from diverse users whose objectives may fundamentally differ or even conflict. This motivates studying methods that can learn from heterogeneous crowd feedback while still enabling alignment to individual users.

\paragraph{Personalized RLHF and Pluralistic Alignment}

Recent work has begun to investigate how to model heterogeneous human preferences and personalize AI behavior to different users. Several approaches have explored preference diversity primarily in the context of large language models (LLMs)~\citep{pca,Pref_pulkit,dpl_siththaranjan2024distributional}. These methods typically model preference variation through latent variables or preference decompositions, but focus on reward modeling rather than policy personalization or embodied control domains.
Distributional Preference Learning (DPL)~\citep{dpl_siththaranjan2024distributional} studies preference learning in the presence of \emph{hidden context}, where annotator identity or other contextual factors influencing preferences are not observed by the reward model. 
% The authors show that standard preference learning implicitly aggregates preferences across such hidden contexts via a Borda-count–like mechanism and propose modeling reward outputs as distributions to detect and mitigate these effects. 
While this approach explicitly models preference heterogeneity through latent contextual factors during reward learning, its primary goal is to characterize and detect hidden context rather than to construct personalized policies for new users.
Only a small number of works study pluralistic alignment in embodied control tasks. Variational Preference Learning (VPL)~\citep{vpl_poddar2024personalizing} models user-specific reward functions by introducing a latent variable conditioned on annotator identity, requiring access to annotator IDs during training. Alternatively, \citet{pareto_learning} learns a Pareto set of reward hypotheses using lexicase selection to represent trade-offs among multiple objectives.
Another line of prior work \cite{qian2026icml_implicit} identifies shared safety principles from crowd feedback but does not consider personalization or model user differences as we do.
Despite many recent advances in pluralistic alignment, existing approaches~\cite{dorsa-myers22a,vpl_poddar2024personalizing,pareto_learning} either require annotator identities, explicitly model latent contexts during reward learning, or produce sets of reward hypotheses rather than directly enabling personalization for new users or learn shared skills from crowd feedback rather than personalizing for individual preferred behaviors.
In contrast, our framework learns a shared basis of preference directions from anonymized heterogeneous crowd feedback. This basis representation allows personalized reward functions—and corresponding policies—to be constructed for previously unseen users at test time from a small number of preference queries.
\begin{figure*}[t]
     \centering
         \includegraphics[width=0.85\linewidth]{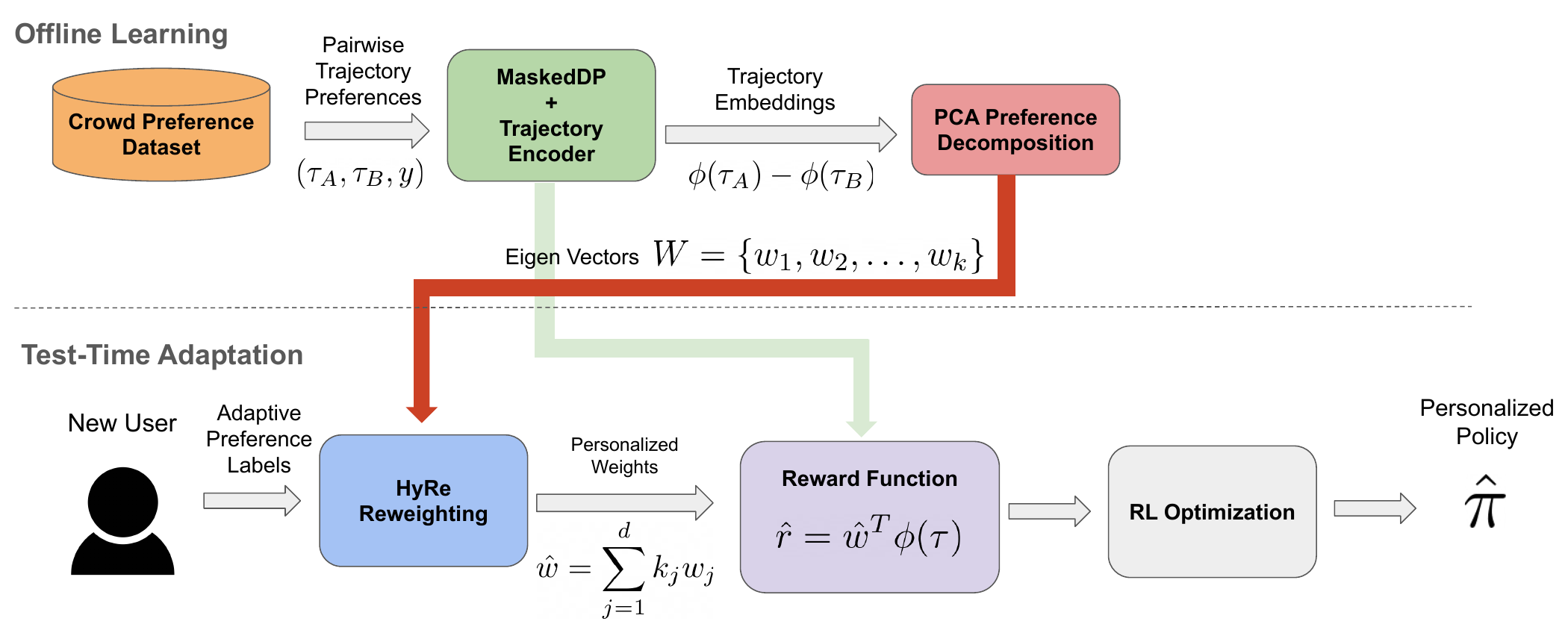}
        
         \caption{\textbf{Preference-PCA Pluralistic Learning (P3L)}}
        \label{fig:pipeline-P3L}
\end{figure*}

\section{Preliminaries}

\paragraph{RLHF}
We model the problem as a Markov Decision Process (MDP)~\citep{mdp_book} with state space $S$, action space $A$, horizon $T$, and reward function $r : S \times A \rightarrow \mathbb{R}$. The true reward function $r$ is unknown and must be inferred from human preferences over trajectories. 
A trajectory $\tau$ is defined as a sequence of state–action pairs over the horizon $T$, expressed as $\tau = (s_1, a_1,  \dots, s_{T-1}, a_{T-1})$.
% COMMENT FOR RLBREW
% \[
% \tau = (s_1, a_1, s_2, a_2, \dots, s_{T-1}, a_{T-1}).
% \]
Given a dataset $D$ of annotated preferences over trajectory pairs of the form $\tau_i \prec \tau_j$, traditional RLHF first learns a reward function using a maximum likelihood objective based on the Bradley–Terry–Luce (BTL) model~\citep{bradley1952rank}, written as $p(\tau_i \succ \tau_j) =
\frac{e^{r_\theta(\tau_i)}}{e^{r_\theta(\tau_i)} + e^{r_\theta(\tau_j)}}$,
where $\theta$ denotes the parameters of the learned reward model.
A policy $\pi^\star_{r_\theta}$ is then optimized under the learned reward: $\pi^\star_{r_\theta} =
\arg\max_{\pi}
\mathbb{E}_{\pi}
\left[
\sum_{t=0}^{T} \gamma^{t} r_{\theta}(s_t,a_t)
\right]$.
This formulation learns a \emph{single} reward function that aggregates all preferences. Consequently, divergent preference signals are treated as noise rather than as evidence of different user objectives.

\paragraph{\textbf{Variational Preference Learning (VPL)}} \label{vpl_limitation}
% \textcolor{red}{Quick note- ASK Prof- should we talk about contextual bandit here and how do we argue that it's weaker with respect to horzion ie VPL used only 1 timestep/only 1 choice wrt object even in Habitat? }
Prior work by~\citet{vpl_poddar2024personalizing} addresses learning from crowd preferences by introducing a latent preference variable $z$ associated with each annotator. The reward model is then conditioned on this latent variable via a contextual BTL model $p(\tau_i \succ \tau_j \mid z) =
\frac{e^{r_\theta(\tau_i,z)}}{e^{r_\theta(\tau_i,z)} + e^{r_\theta(\tau_j,z)}} .$
% COMMENT FOR RLBREW
% \begin{equation}
% p(\tau_i \succ \tau_j \mid z) =
% \frac{e^{r_\theta(\tau_i,z)}}{e^{r_\theta(\tau_i,z)} + e^{r_\theta(\tau_j,z)}} .
% \end{equation}
However, this formulation assumes access to annotator identities during training in order to infer the latent preference variable.
VPL also only considers preference elicitation in one-step, contextual bandit~\citep{vpl_poddar2024personalizing} settings. 
In contrast, the longer-horizon embodied control tasks we consider require  distinguishing between temporally extended trajectories and their cumulative consequences. 
Another limitation of VPL that we address is that it requires a fixed number of preference queries for adaptation to a new user, referred to as the \emph{context set}~\citep{vpl_poddar2024personalizing}. This constraint arises from the VAE-based latent preference inference architecture, whose encoder requires a fixed-size input. 
Because each query must be labeled by the new user, this requirement introduces additional  interaction and annotation cost, and precludes test-time learning from variable numbers of adaptation queries, something that our approach enables.
Finally, prior work has shown that VPL is susceptible to posterior collapse~\cite{kim2026swapguided_iclr}.
%\citet{kim2026swapguided_iclr} identify two complementary causes of posterior collapse in VPL: (1) complex and potentially noisy preference feedback can destabilize the learning of latent user representations and (2) a sufficiently expressive decoder can bypass this representation altogether by predicting preferences without conditioning meaningfully on z.
% In this regime, the user-specific latent variable becomes uninformative and is effectively ignored by the reward model, causing VPL to revert toward an implicit single-reward model and undermining personalization. 
We find that this limitation extends beyond language modeling to embodied control, where we observe collapse in large state-spaces and longer-horizon tasks. For more details, please refer to Appendix ~\ref{app:vpl_posterior_collapse}.

\paragraph{Decomposed Reward Models (DRM)}

\citet{pca} reformulate preference learning by decomposing rewards into multiple orthogonal preference directions in representation space rather than learning a single scalar reward. 
% This decomposition enables modeling diverse user preferences through linear combinations of these directions.
However, their analysis focuses primarily on reward prediction accuracy and does not consider optimizing policies under personalized reward functions. High accuracy on held-out preference queries does not necessarily translate to behaviorally aligned policies, as widely documented in prior RLHF work~\citep{christiano2017deep,tien2023causal,wen2025rethinking}.
Furthermore, DRM only tested this idea on pretrained LLM embedding models trained on internet-scale data such as Gemma-2B-RM and Llama3-8B-RM, which have undergone multiple stages of training including pretraining and supervised instruction fine-tuning before preference alignment~\citep{ziegler2020finetuninglanguagemodelshuman,ouyang2022training}. 
% \citet{pca} show that these pretrained models provide rich representations capable of capturing diverse human preference factors.
In contrast, many embodied control domains lack such pretrained representation models. A key challenge that we tackle in this paper is to learn meaningful trajectory representations  and preference decompositions  directly from limited task-specific data in settings where large pretrained embedding models are unavailable.
% Another significant challenge is  learning the joint representation of user distribution as well as task objectives , which is essentially harder than just learning user preference distribution on supervised finetuned embeddings by LLMS, as supervised finetuning disentangles learning task objectives first and then user objectives are learned by RLHF in final stage. 
%While supervised fine-tuning first anchors LLMs to the desired task behavior, so RLHF can focus on modeling human preferences, 
We show that it is possible to jointly learn the task objectives as well as user preference distribution from heterogeneous preference dataset using only self-supervised representations.
% In our setting, this decomposition is unavailable: task objectives and the user preference distribution must be learned jointly from self-supervised representations.
\subsection{Problem Formulation}

Learning from diverse crowdsourced feedback presents a more challenging setting than standard RLHF. 
Vanilla RLHF aggregates preferences from multiple annotators into a single reward function, implicitly assuming that all preference labels arise from a shared underlying objective. However, in many real-world settings different users may have fundamentally different goals or behavioral preferences. 
The idea of creating AI systems that can model and serve pluralistic human values is known as \emph{pluralistic alignment}~\citep{roadmap_to_pl_align}.
% Our work specifically focuses and delves into steerable pluralistic alignment, aimed at  steering (aligning) an AI system to the preferences of a specific user at test time in the presence of heterogeneous crowd feedback is known as \emph{ Steerably pluralistic alignment}~\citep{roadmap_to_pl_align}.
Our work specifically focuses on \emph{steerable pluralistic alignment}, where the goal is to adapt an AI system at test time to the preferences of a specific user in the presence of heterogeneous crowd feedback.

\paragraph{Steerable Pluralistic Alignment with Heterogeneous Users}
Consider a population of $N$ annotators, where each user $u_i$ has an unknown ground-truth reward function $r_i$. For each preference query $m$, an unobserved annotator index $z^m \in \{1,\dots,N\}$ generates a label $y^m \in \{0,1\}$ over a trajectory pair $(\tau_A^m,\tau_B^m)$ according to that annotator's latent reward function $r_{z^m}$, where $y^m = 0$ indicates that $\tau_A^m$ is preferred over $\tau_B^m$. Because users may have different objectives, conflicting labels for the same pair of trajectories can naturally occur
% , e.g.,
% $(\tau_A,\tau_B,y=0) \quad \text{and} \quad (\tau_A,\tau_B,y=1)$,
% \[
% (\tau_A,\tau_B,y=0) \quad \text{and} \quad (\tau_A,\tau_B,y=1),
% \]
when the two labels are produced by different annotators.
We assume access to an offline dataset of crowd preferences $
D = \{(\tau_A^m,\tau_B^m,y^m)\}_{m=1}^{M}$,
% \[
% D = \{(\tau_A^m,\tau_B^m,y^m)\}_{m=1}^{M},
% \]
where the annotator identities $\{z^m\}_{m=1}^M$ may or may not be available.

\paragraph{Learning Objective}

Our goal is to use this heterogeneous preference dataset to learn representations of heterogeneous preferences that enable efficient test-time personalization of reward functions and policies for new users. 
% In the primary setting, we do not assume access to annotator identities during training. 
% However, if annotator information is available, the framework should be able to leverage it to improve representation learning and alignment.
Under this setting, we aim to address the following questions:
% \begin{enumerate}
% \item Can we infer personalized reward functions $\hat r_i$ for individual users from heterogeneous preference data without requiring annotator identities and without any known goal objectives?
(1) Can we infer personalized reward functions $\hat r_i$ for individual users from heterogeneous preference data when neither annotator identities nor the underlying task objectives are known?
(2) Can we optimize policies $\hat{\pi}_i$ under these inferred rewards to produce behaviors aligned with diverse user preferences?
(3) Can such personalization generalize to both in-distribution users and to users whose rewards fall outside the distribution of the %previously unseen users relative to the 
offline preference dataset?
% \end{enumerate}

\section{Preference-PCA Pluralistic Learning (P3L)}
We term our proposed approach Preference-PCA Pluralistic Learning (P3L). The overall pipeline of our approach is shown in Fig.~\ref{fig:pipeline-P3L}.
We consider the joint problem of pre-training an expressive embedding extractor  and leveraging it to learn  preference vectors from diverse crowd preferences in control domains to  provide personalized alignment to test-time users on both test time held out set as well as a personalized policy optimized under the personalized derived reward.

\subsection{Representation Encoders}
We study two types of representation encoders that employ Masked Decision Prediction (MaskedDP) ~\citep{maskeddp} as their base architecture, and describe each below.

\subsubsection{Anonymized CLS-MaskedDP:}
In this setting, we assume anonymized crowd data with no knowledge about annotator ids. 
We learn the trajectory embedding following the CLS-token formulation of BERT~\citep{devlin2019bert}, by feeding the outputs of frozen MaskedDP encoder through a self-attention module, thereby learning a CLS token in a completely end-to-end unsupervised manner. 
We learn the summary token by employing different masking ratios in MaskedDP in the triplet loss. 
Given two trajectories $\tau_{1},\tau_{2}$ from the preference dataset, we use $\phi_{\tau_{1}}$ with mask ratio $m_{1}$ for anchor and with another mask ratio $m_2$ as the positive sample, while $\phi_{\tau_{2}}$ with mask $m_{1}$
 is used as a negative sample 
%  \begin{equation}
% \mathcal{L} 
% = \max \left( 0,\; 
% \| \phi((\tau_{1})_{m1}) - \phi((\tau_{1})_{m2}) \|_2^2 
% - 
% \| \phi((\tau_{1})_{m1}) -\phi((\tau_{2})_{m1}) \|_2^2 
% + \alpha 
% \right)
% \end{equation}
\begin{equation}
\begin{aligned}
\mathcal{L}
= \max \Bigl(0,\;
&\| \phi(\tau_{1}|m_1) - \phi(\tau_{1}|m_2) \|_2^2 \\
&- \| \phi(\tau_{1}|m_1) - \phi(\tau_{2}|m_1) \|_2^2
+ \alpha
\Bigr),
\end{aligned}
\end{equation}
where $\alpha$ is the triplet margin. 
This approach is a completely end-to-end unsupervised embedding learning method and we refer to this approach later on as \textbf{CLS-MaskedDP}.

\subsubsection{Privileged Traj-MaskedDP:}
In this setting, we assume that annotator id is known. 
This is the exact same assumption made in VPL by ~\citet{vpl_poddar2024personalizing}. We term this approach as \textit{privileged} because of this assumption about access to privileged information about annotator identity.
% We leverage knowledge about different annotators to learn the linear summarization using triplet loss.
We leverage prior knowledge about annotator-identity on preference  labels to learn a linear summarization of the trajectory. 
% Given annotator-identity, we sample 2 trajectories from one annotator$A_1$ set to serve as $\tau_{anchor}$ and $\tau_{pos}$ while a trajectory from another annotator$A_2$ which acts as $\tau_{neg}$.
Formally, given a preference dataset D=\{$(\tau_b,\tau_c), {(\tau_d,\tau_e}),...(\tau_o,\tau_m)$\}  of size M with knowledge that annotator $A_1$ prefers $\tau_b \succ \tau_c$ and $\tau_d \succ \tau_e$, we sample $\tau_b$ and $\tau_d$ from $A_1$'s set to serve as the anchor and positive trajectories respectively while  for another annotator $A_2$ given that $\tau_o \succ \tau_m$, then we can use $\tau_o$  as the negative trajectory.
Latent embeddings generated by frozen MaskedDP encoder are fed into a linear layer trained using contrastive learning via a triplet loss:
\begin{equation}
\begin{aligned}
\mathcal{L}
= \max \Bigl(0,\;
&\|\phi(\tau^{A_1}_{anchor})-\phi(\tau^{A_1}_{pos})\|_2^2 \\
&-\|\phi(\tau^{A_1}_{anchor})-\phi(\tau^{A_2}_{neg})\|_2^2
+\alpha
\Bigr)
\end{aligned}
\end{equation}
where $\alpha$ is the triplet margin. 
% Note, 
We refer to this approach as  \textbf{Traj-MaskedDP}.
% We feed the latent embedding outputed from frozen MaskedDP encoder for each of these trajectories to a linear layer to train it using contrastive learning via triplet loss . Formally, 

% Given a finite set of H objectives with each annotator preferring one specific objective where $\mathcal{H} = \{ h_1, h_2, \dots,h_H\}$ , we sample 2 objectives i.e. $h_{a} \neq h_{b}$.
% We then use first objective $h_{a}$to sample 2 trajectories to form an anchor $\tau_{anchor}$ and a positive $\tau_{pos}$ and sample a negative $\tau_{neg}$ from 2nd objective $h_{b}$ to learn a single trajectory embedding via triplet loss where $\alpha$ is the 

% \begin{equation}
% \mathcal{L} 
% = \max \left( 0,\; 
% \| \phi(\tau^{A_1}_{anchor}) - \phi(\tau^{A_1}_{pos}) \|_2^2 
% - 
% \| \phi(\tau^{A_1}_{anchor}) - \phi(\tau^{A_2}_{neg}) \|_2^2 
% + \alpha 
% \right)
% \end{equation}

\subsection{Learning from Crowd-Sourced Feedback}
 Given the crowd-sourced preference dataset $D$ with preferences over trajectories of form $\tau_i \prec \tau_j$, we model  heterogeneity in human choices  as a preference vector via the following steps.

\paragraph{Learning eigenvectors}
We first apply Principal Component Analysis (PCA)~\cite{pearson1901lines} to a set of latent difference embeddings $\Phi_\text{diff}$ derived from pairwise preference queries in the offline heterogeneous preference dataset.
This yields a  set of principal basis vectors $W = [w_1, w_2, . . . , w_d]$, where  each orthogonal basis vector captures a principal direction of variation in the observed preference signals. The individual latent difference embeddings are defined as $\phi_{diff} =\phi(\tau_1) - \phi(\tau_2) \text{ for } y=0$ and $\phi_{diff} =\phi(\tau_2) - \phi(\tau_1) \text{ for } y=1$.
Similar to DRM~\cite{pca}, we assume that human preferences can be modeled as approximately linearly decomposable along orthogonal directions, $W$ in the representation space, where
\begin{equation}
\begin{aligned}
W &= \{w_i\}_{i=1}^{k}
    = \mathrm{PCA}(\Phi_{\text{diff}}). \\
% \phi_{\text{diff}} &=
% \begin{cases}
% \phi(\tau_1)-\phi(\tau_2), & y=0, \\
% \phi(\tau_2)-\phi(\tau_1), & y=1.
% \end{cases}
\end{aligned}
\end{equation}
% Both positive $w_{i}$  and negative $-w_{i}$ of the eigenvector are taken to account for fact that PCA eigenvectors are direction agnostic while human preference is inherently  not.  
% Both $w_{i}$ and $-w_{i}$ are retained to account for the inherent sign ambiguity of PCA eigenvectors. PCA identifies an axis of variation, but does not assign a meaningful positive or negative orientation to that axis.
Since PCA is invariant to eigenvector sign, the direction $w_{i}$is equivalent to $-w_{i}$ from the perspective of variance decomposition.
In contrast, human preferences are directional: moving along $w_{i}$
 may correspond to favoring one behavioral tendency, while moving along $-w_{i}$ may correspond to the opposing tendency. 
We therefore preserve both orientations $w_{i}$ and $-w_{i}$  of each principal component to capture contrasting behavioral preferences.
 % Considering both signs therefore preserves the two possible preference directions associated with each principal component.
Thus any human preference can be modeled as linear combination of these basis vector $\hat{w}=\sum_{j=1}^{d} k_{j}w_{j} $.
where $k_1,k_2,...k_d$ are weights on these individual vectors $w_{i}$. 
% \begin{equation}
%   \hat{w}=\sum_{j=1}^{d} k_{j}w_{j} .
% \end{equation}
% where $k_1,k_2,...k_d$ are weights on these individual vectors $w_{i}$. 
% Each $w_{i}$ can be combined with the embedding $\phi(\tau)$ to give the reward of trajectory $\tau$ in that dimension and 
The total predicted reward of trajectory $\tau$ can be expressed as $\hat{r}=\hat{w}^T\phi(\tau)$.
% \begin{equation}
%     \hat{r}=\hat{w}^T\phi(\tau).
% \end{equation}

\paragraph{Test-Time User Adaptation} %Adapting to Test time User}
At test time, we actively query the user for their preferences to form an adaptive set of queries $D_{adapt}$.
We then leverage these adaptive queries to compute coefficients $k_1,k_2,...k_d$ on the principal eigenvectors using Hypothesis Reweighing (HyRe) ~\citep{HyRe} to form the complete preference vector of the test-time user: \vspace{-6mm}
\begin{center}
\footnotesize
\[
\begin{aligned}
k_i
% &=
% \mathrm{softmax}\!\left(
% \mathbb{E}_{(\tau_A,\tau_B,y)\sim D_{\text{adapt}}}
% \left[
% \log \sigma\!\Big(
% \hat{r}_i(\tau_A) - \hat{r}_i(\tau_B)
% \Big)
% \right]
% \right) \\
% &=
% \frac{
% \exp\!\left(
% \mathbb{E}_{(\tau_A,\tau_B,y)\sim D_{\text{adapt}}}
% \left[
% \log \sigma\!\Big(
% R_i(\tau_A) - R_i(\tau_B)
% \Big)
% \right]
% \right)
% }{
% \sum_{d}
% \exp\!\left(
% \mathbb{E}_{(\tau_A,\tau_B,y)\sim D_{\text{adapt}}}
% \left[
% \log \sigma\!\Big(
% R_d(\tau_A) - R_d(\tau_B)
% \Big)
% \right]
% \right)
% } \\
&=
\frac{
\exp\!\left(
\mathbb{E}_{(\tau_A,\tau_B,y)\sim D_{\text{adapt}}}
\left[
\log \sigma\!\Big(
w_i^\top
\big(\phi(\tau_A)-\phi(\tau_B)\big)
\Big)
\right]
\right)
}{
\sum_{d}
\exp\!\left(
\mathbb{E}_{(\tau_A,\tau_B,y)\sim D_{\text{adapt}}}
\left[
\log \sigma\!\Big(w_d^\top
\big(\phi(\tau_A)-\phi(\tau_B)\big)
\Big)
\right]
\right)
}
\end{aligned}
\]
\end{center}
In this manner, weights $k_j$ shift dynamically according to the adaptive preference dataset to model the new user.
% by assigning greater weight to directions that capture more of this preference direction by assigning higher weight to eigenvectors $w_j$ that minimize the loss while down-weighting those associated with higher loss. 

\paragraph{Learning a Personalized Policy}
Once we have the weights $k_j$ on our basis eigenvectors $w_j$, we then learn a personalized policy $\hat{\pi}$ for the new user $i$ by using reinforcement learning to maximize the expected performance under $\hat{r_i}$. 
Formally $\hat{\pi}_i^{\star}
=
\arg\max_{\pi_i}
\mathbb{E}_{\tau \sim \pi_i}\!\left[\hat{r}_i(\tau)\right]
=
\arg\max_{\pi_i}
\mathbb{E}_{\tau \sim \pi_i}\!\left[\hat{w}_i^{\top}\phi(\tau)\right]$.

% \begin{equation}
% \hat{\pi}_i^{\star}
% =
% \arg\max_{\pi_i}
% \mathbb{E}_{\tau \sim \pi_i}\!\left[\hat{r}_i(\tau)\right]
% =
% \arg\max_{\pi_i}
% \mathbb{E}_{\tau \sim \pi_i}\!\left[\hat{w}_i^{\top}\phi(\tau)\right]
% \end{equation}.
\begin{figure}[t]
     \centering
         \includegraphics[width=\linewidth]{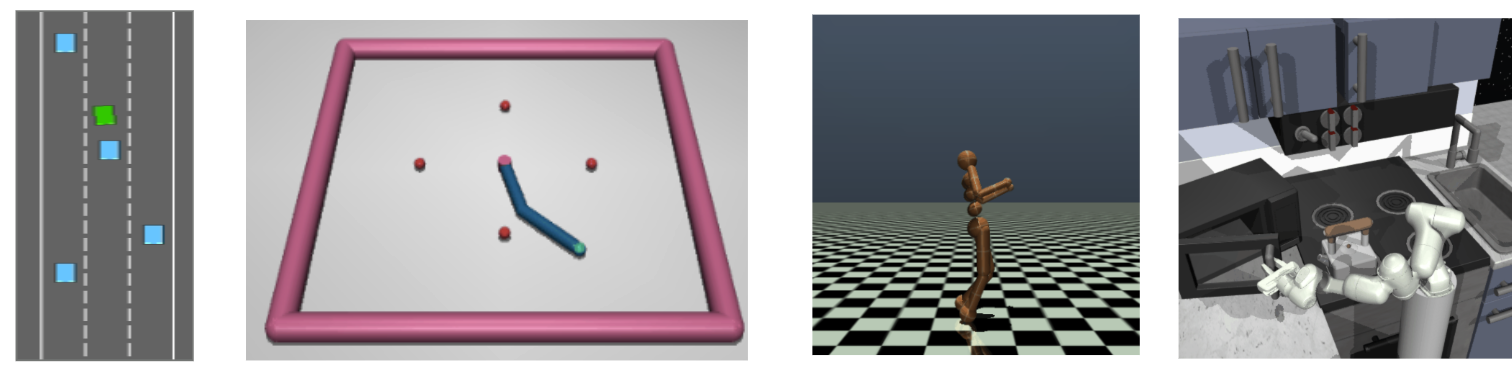}
         \caption{\textbf{Environments} MO-Highway, MO-Reacher, MO-Humanoid and MO-Kitchen.}
        \label{fig:Envs}
        
\end{figure}
\section{Experimental Setup}
We first describe the environments used, followed by the baseline methods for comparison and finally present the evaluation criteria and metrics. 
% We first describe the enviornments considered , then briefly discuss baselines compared against and finally discuss the metrics used. 
% Ask about order of envs - Highway then Reacher then Humanoid and then Franka or Reacher ,Humanoid , Highway and then Franka 

% Our experimental setup is as follows : 
\paragraph{Environments}
Following the pluralism evaluation benchmark criteria outlined in  ~\citet{roadmap_to_pl_align}, we consider multi-objective environments for steerable pluralistic alignment. 
% Motivated by the pluralism benchmark criteria outlined by~\citet{roadmap_to_pl_align}, we consider multi-objective environments for evaluating steerable pluralistic alignment.
They provide a principled way to represent crowdsourced feedback with diverse and potentially conflicting objectives. Fig.~\ref{fig:Envs} shows the environments used: MO-Highway and MO-Reacher from MO-Gymnasium~\cite{MORL_benchmarking}, and MO-Humanoid and MO-Kitchen, described below.
% To evaluate steerable pluralistic alignment, we use multi-objective environments as recommended by ~\citet{roadmap_to_pl_align}. 
% To learn from crowdsourced feedback with multiple—possibly conflicting—objectives, we consider the environments depicted in Fig ~\ref{fig:Envs}.
% MO-Highway and MO-Reacher are drawn from MO-Gymnasium~\cite{MORL_benchmarking}, while MO-Humanoid and Franka-Kitchen are described below.
% where MO-Reacher and MO-Highway are from MO-Gymnasium~\cite{MORL_benchmarking}, and MO-Humanoid and Franka-Kitchen described below.
% from MO-Gymnasium as described by ~\citet{MORL_benchmarking}:

\textbf{MO-Highway} has a 3-dimensional reward with first reward dimension rewarding high-speed driving while the second dimension scores on driving in right lane and the last reward dimension represents the collision reward. 

\textbf{MO-Reacher} provides a 4-dimensional reward vector, where each dimension corresponds to the reward for reaching a specific goal.
Some goals induce competing preferences; for example, reaching the left goal conflicts directly with reaching the right goal.
% has 4d reward vector where each distinct dimension represents  the reward achieved by agent for one particular goal. 
% Note that some of these goals yield competing preferences in the preference dataset as reaching goal on left is completely conflicting to reaching the goal on right. 
% ii) MO-FourRooms iii) 
% iii) \citet{reward-free_mo_iclr26} introduced a 5d MO-Humanoid  with five objectives: maximizing moving speed along the x-axis, maximizing moving speed along the y-axis, maximizing angular velocity on the left elbow,
% maximizing angular velocity on the right elbow, and minimizing energy cost. five objectives: maximizing moving speed along the x-axis, maximizing moving speed along the y-axis, maximizing angular velocity on the left elbow,
% maximizing angular velocity on the right elbow, and minimizing energy cost. It consists of a high dimensional continuous state space  376 and an action space of 17

\textbf{MO-Humanoid}~\cite{reward-free_mo_iclr26} has a 5-dimensional objective space with objectives corresponding
to forward velocity along the $x$-axis, lateral velocity along the
$y$-axis, angular velocities of the left and right elbows, and energy
efficiency. 
The environment has a high-dimensional continuous
observation space of dimension 376 and an action space of dimension 17.
% iv)Franka-Kitchen~\cite{franka_kitchen-gupta20a}, is a popular robotic benchmark that features only a 9-DoF Franka robot in a complex kitchen environment where different users might prefer different tasks. It represents a harder contact-rich robotic manipulation task where the robot needs to interact with different objectives and perform completely varied tasks from opening the microwave door to pushing the kettle to turning oven knobs for different burners.

\textbf{MO-Kitchen} is a multi-objective version of Franka Kitchen~\cite{franka_kitchen-gupta20a} %is a contact-rich robotic manipulation benchmark from Gymansium Robotics~\cite{gymnasium_robotics2023github} 
that we created,
featuring a 9-DoF Franka robot that can perform several distinct manipulation tasks, such as opening the microwave, moving the kettle, and turning oven knobs for different burners. 
This setting naturally induces heterogeneous user preferences such as User 1 might prefer opening the microwave, User 2 moving the kettle, User 3 toggling the light switch, while  User 4 prefers turning the knob to activate the top burner.
% Thus we engineered MO-Kitchen to have multi-objective vectorized reward to capture all possible tasks that a user can prefer. 
% These heterogeneous tasks provide a natural setting for modeling user preferences, where different users may
% prefer different subsets of tasks or objectives.
% Franka-Kitchen is also substantially more challenging than the simpler control environments because it requires long-horizon, contact-rich manipulation, where
% successful behavior depends on extended sequences of coordinated actions
% and precise physical interactions. 
MO-Kitchen is challenging because feedback from multiple users with different preferred tasks is learned jointly, requiring the framework to disentangle heterogeneous user objectives 
% heterogeneous preferences over distinct long-horizon goals where motion to the object, contact with the object as well as orientation all play a significant role in successfully achieving a task .
while reasoning over coordinated reaching, contact-rich manipulation, and precise end-effector orientation.

For MO-Kitchen, we conduct experiments and explore personalization in multiple variants, starting from personalizing in simple setting with only 2 possible tasks, termed as \emph{MO-Kitchen-2d} (opening the microwave door or moving the kettle) and then progress to challenging scenarios with 3 possible tasks, termed as \emph{MO-Kitchen-3d} (microwave door, kettle, or flipping the light switch). Finally we move to the most sophisticated domain with 4  different tasks possible,  termed as \emph{MO-Kitchen-4d},  where each user can choose from a variety of tasks from opening the microwave door to pushing the kettle to turning oven knobs for different burners to turning on the light switch.

% ---or
% ii) successful execution of a task requires precise reaching, contact-rich manipulation, and orientation control.
% ---or
% iii) over distinct long-horizon goals where motion to the object, contact with the object as well as orientation all play a significant role in successfully achieving a task 

% As a result, preference-relevant differences may only become apparent over longer trajectories, making preference learning more challenging.

% These heterogeneous tasks provide a natural setting for modeling user preferences, where different users may prefer different subsets of tasks or objectives.
% Franka-Kitchen is also substantially more challenging than the simpler control environments because it requires long-horizon, contact-rich manipulation: successful behavior often depends on executing extended sequences of coordinated actions, maintaining precise control through physical interactions, and handling dependencies between intermediate actions before the desired task outcome is achieved. 
% This makes preference learning harder, since the consequences of a user’s preference may only become apparent after many timesteps.

In each of these environments we ignore the terminal condition and make these tasks fixed time horizon to avoid leakage of information and to prevent termination bias in reward learning from preferences as advised by ~\citet{freire2020derail}.
Based on the number of underlying objectives in an environment we report the performance across all objectives where each individual user prefers one specific objective.
For simplicity in our experiments, we associate User 1 with a preference for Objective 1 over all other objectives, User 2 with Objective 2, and so on. 
This is only a notational convention; our framework  makes no assumption itself that user and objective indices must align, and thus in general, any user may prefer any objective.
% To keep it straightforward for our experiments, we consider User 1 represents preferring objective 1 over any other objective and so on. Our framework makes no assumption about 1st user preferring 1st objective and any user is completely free to prefer any objective they like.  

\paragraph{Baselines}
We evaluate both our privileged and non-privileged approach against a single reward head model (SingleRM) learned through conventional Bradley-Terry loss optimization, and VPL. 
% Originally VPL was trained with input as only state, and for consistency with original work, we keep the input as state only. 
VPL uses adaptive queries to infer the latent preference variable on which it conditions its reward and policy while our approaches, inspired by~\citep{pca} use HyRe~\citep{HyRe}, to infer the test-time user's preference vector. 
For complete setup details with respect to number of queries, etc., please refer to Appendix~\ref{app:preference_setup_details}.
% We evaluate both our privileged and non-privileged approach against 1) Single Reward Head Model learned through conventional Bradley  Terry loss Model, 2) VPL - 2 versions of VPL as original VPL was trained with input as only state , we term it as VPL-v1 . 
% To have a more rigorous comparison of VPL to our approach, we also train a VPL using both state and action as inputs , we term it as VPL-v2.
We optimize online RL policies under learned reward functions using Augmented Random Search ~\citep{ars} and Proximal Policy Optimization (PPO) ~\citep{schulman2017proximal} from Stable Baselines3 ~\citep{sb3} with episodic rewards.
We employ episodic rewards for two reasons: (1) because the preference dataset is constructed using fixed horizon trajectories and we want  to mitigate termination bias~\citep{freire2020derail} and (2) because
our reward formulation combines the PCA preference vector with a representation encoder that summarizes the full trajectory. 
% Since Franka-Kitchen was released for offline RL, for all frameworks, we optimize a continuous IQL policy under learned reward but evaluate it in online settings. 

Original VPL is an offline framework which uses Implicit Q-Learning (IQL) \citep{kostrikov2022offline} to learn an offline policy optimized for per timestep reward. 
% From TorchRL~\citep{bou2024torchrl}, we use discrete IQL for MO-Highway and MO-Reacher and continuous IQL for MO-Humanoid and Franka-Kitchen respectively.
For fair comparison to our frameworks, we learn both an offline IQL policy as well as an online episodic policy for VPL for all MO-environments.
Since MO-Kitchen is commonly used as an offline RL benchmark, we optimize a continuous-control IQL policy on the collected offline dataset using the learned personalized reward, and evaluate the resulting policy through online environment rollouts.
For more details on RL optimization and hyper-parameters, see Appendix~\ref{app:RL_exp_details}.

% since reward is computed using PCA preference vector and our representation encoder that summarizes the whole trajectory ~\ (see Appendix ~\ref{app:RL_exp_details} for full details).

% We also show in Appendix{ADD REF LATER} that ARS outperforms PPO when used to optimize policy under episodic settings{Should we mention tasks themselves are not episodic but we use MaskedDP is used as frozen encoder before trajectory summarization }. 
% Prior work~\citep{pareto_learning} also uses gradient free optimization methods for personal alignment to the test-time user. 

% under each of the VPL reward models VPL-v1 and VPL-v2. 

\paragraph{Metrics}
We compare the performance of our approach against earlier frameworks on these metrics: 
(1) accuracy on adapting to a new held out test-time user's preferences;
(2) performance of RL policy optimized under new test-time user's preferences;
(3) time taken on NVIDIA GeForce RTX-GT3090 to train these models;
(4) Ablation with respect to the number of adaptive queries (For ablation with respect to training queries and number of PCA dimensions, see Appendix ~\ref{app:num_training_queries_ablation} and Appendix ~\ref{app:pca_dim_ablation} respectively); and
(5) Adaptation to out-of-distribution users where we train on subset of different users and test the adaptation to held-out new user both on test and RL policy.

% We report metrics (3-5) on MO-Reacher here a

 % Results for the other environments are provided in  Appendix~\ref{app:additional_exps_more_envs} for metrics (3) and (4), and in Appendix~\ref{app:more_ood_eval} for metric (5).
Note that in the Original VPL paper~\citep{vpl_poddar2024personalizing}, the adaptation query set size is referred to as context size in VPL and is noted in their appendix as the main hyperparameter to tune. 
Therefore, metric (4) directly evaluates prior work under its key adaptation hyperparameter, and we include this ablation to ensure a fair comparison against VPL.
 % We report metrics (3)–(5) for MO-Reacher here 
 % and for more results across all other environments please refer to Appendix ~\ref{app:additional_exps_more_envs} for metrics (3) and (4) and Appendix ~\ref{app:more_ood_eval} for metric (5).

\begin{table*}[t]
\centering
% \footnotesize
\scriptsize
\setlength{\tabcolsep}{9pt}
  
  \centering
  \begin{tabular}{lrrrr}
 & \multicolumn{2}{c}{Baselines} & \multicolumn{2}{c}{P3L} \\
 \cmidrule(r){2-3} \cmidrule(lr){4-5}
    Environment  & 
    SingleRM    & VPL  & Privileged-Traj-MaskedDP  & Anonymized-CLS-MaskedDP \\
    \midrule
    MO-Highway  & 70.222 ± 3.326 & 66.667 ± 3.143 & \best{84.889 ± 5.145} &\second{78.667 ± 4.576}  \\
\midrule
MO-Reacher & 54.667 ± 0.667 & 52.00 ±3.055 & \second{93.000 ± 1.453} & \best{93.500 ± 1.434} \\
\midrule
MO-Humanoid & 76.280 ± 1.294 & 74.920 ± 2.018 & \best{88.320 ± 0.515} & \second{77.880 ± 1.017}\\
\midrule
MO-Kitchen-2d & 50.000 ± 0.000
 &  50.400 ± 0.800 & \best{100.000 ± 0.0 } & \second{90.400 ± 4.079}\\  
       \midrule
          MO-Kitchen-3d &53.185 ± 1.582 & 51.407 ± 1.525 & \best{84.148 ± 2.227 } & \second{82.667 ± 3.647} \\
          \midrule
MO-Kitchen-4d & 
  55.278 ± 2.159& 51.722 ± 2.905 &\best{78.444 ± 0.572} & \second{72.167 ± 2.021}\\
\bottomrule
  \end{tabular} 
    \caption{\textbf{ Average user accuracy of held out test-set preference queries on different environments reported as Mean and Std across 5 seeds with number of different users possible on different objectives reflected next to each environment}. Both of our approaches, \textbf{Privileged-Traj-MaskedDP} which assumes privileged access to annotator id (as does VPL), as well as \textbf{Anonymized-CLS-MaskedDP} which does not assume any access to annotator identity, outperform both the  baselines Single Reward Model and VPL on average across all users. The blue background indicates the best performance, while the gray background denotes the second-best performance.}
    \label{All-env-Average-Acc-table}
\end{table*}

\section{Results}
\subsection{In-Distribution Experiments}
% \subsubsection{
% Accuracy on adapting to a new held out test time user’s preference}
\paragraph{Test-Time Preference Alignment Accuracy}

In Table~\ref{All-env-Average-Acc-table}, we present the average accuracy of adapting to test-time held out preference queries in terms of mean and standard deviation across 5 seeds for all environments. 
For the individual user accuracy of adapting to test-time held out preference queries across all environments and for more details, please refer to Appendix ~\ref{app:all-env-per-user-test-acc}.

% please refer to Appendix ~\ref{app:Indv-Highway-Test-Acc}  for MO-Highway and Appendix ~\ref{app:Indv-Reacher-Test-Acc} for MO-Reacher  and Appendix ~\ref{app:Indv-humanoid-test-acc} for MO-Humanoid and Appendix ~\ref{app:Indv-FrankaKitchen-Test-Acc} for Franka-Kitchen respectively. 
% In Table~\ref{Reacher-Acc-table} and Table~\ref{Highway-Acc-table} for MO-Reacher and MO-Highway, we present the accuracy of adapting to test-time held out preference queries in terms of mean and standard deviation across 5 seeds. We use 2 PCA eigenvectors across both our frameworks.
% \textcolor{red}{To be updated-For MO-humanoid Table~\ref{Humanoid-Acc-table} and for Franka Kitchen~\ref{Kitchen-Acc-table} respectively represent accuracy of adapating to held out test preferences.- Will update results lines below once I have accuracy ready across all seeds.}
% For detailed analysis with respect to the number of PCA components, please refer to Appendix~\ref{app:pca_dim_ablation}.

We observe that for all environments, our approaches Privileged-Traj-MaskedDP and Anonymized-CLS-MaskedDP outperform both Single Reward Model and VPL, where Privileged-Traj-MaskedDP makes the same assumptions about prior knowledge of annotator identity as VPL while the Anonymized-CLS-MaskedDP approach has no such prior knowledge. 
Across MO-Kitchen, we observe that while P3L outperforms both the baselines, its performance decreases as the number of preference dimensions increases. Privileged-Traj-MaskedDP falls from 100.0\% to 84.15\% to 78.44\%, while Anonymized-CLS-MaskedDP decreases from 90.4\% to 82.67\% to 72.17\%. 
This degradation is consistent with the increasing difficulty of preference inference in a higher-dimensional objective space; nevertheless, both P3L variants continue to considerably outperform the baselines, indicating that the proposed framework better captures the underlying multi-objective preference structure.
% These results demonstrate that our framework achieves better alignment with held-out preference queries when learning from diverse crowd feedback without access to annotator identities.
Thus we show better alignment to the held out queries when learning from diverse crowd feedback with no information about annotator identity.

\paragraph{RL performance under Test-Time Preference Alignment}
\begin{table*}[t!]
\centering
% \footnotesize
\scriptsize
\setlength{\tabcolsep}{3pt}
  \centering
  \begin{tabular}{lrrrrr}
 & \multicolumn{3}{c}{Baselines} & \multicolumn{2}{c}{P3L} \\
 \cmidrule(lr){2-4} \cmidrule(lr){5-6}
  % Environment   & 
  %   SingleRM-OnlineRL    & VPL-OfflineRL  & VPL-OnlineRL   & Traj-MaskedDP-OnlineRL   & CLS-MaskedDP-OnlineRL \\
   Environment   & 
    SingleRM-RL    & VPL-OfflineRL  & VPL-RL   & Privileged-RL   & Anonymized-RL \\
    \midrule
     MO-Highway & 
     4.303 ± 9.059 & \best{8.409 ± 15.063} & 4.354 ± 18.164 &  \second{7.721 ±  16.435} &  5.126 ± 20.505 \\
    \midrule
    MO-Reacher & 16.704 ± 1.999 & 16.090 ± 12.372 & 20.427 ± 1.616 & \best{31.708 ± 4.520} & \second{28.603 ± 1.996} \\ 
    \midrule
    MO-Humanoid &  -48.433 ± 171.307  &  -51.301 ± 171.186 & -46.240 ± 172.485 & \best{-10.509 ± 169.372}& \second{-38.794 ± 173.861} \\ 
    \midrule
       MO-Kitchen-2d &  0.274 ± 0.274
 & \second{0.363 ± 0.037} & \multicolumn{1}{c}{--} &  \best{0.834 ± 0.154}
& 0.334 ± 0.182 \\
       \midrule
          MO-Kitchen-3d 
          &0.181 ± 0.140&  0.134 ± 0.188 & \multicolumn{1}{c}{--} & \best{0.799 ± 0.194} & \second{0.243 ± 0.325} \\
          \midrule
    MO-Kitchen-4d & 0.000 ± 0.000 &  \second{0.103 ± 0.179}  & \multicolumn{1}{c}{--}   & \best{ 0.661 ± 0.232} & 0.046 ± 0.076\\
    \bottomrule
    
\end{tabular}
 \caption{\textbf{Average RL performance of personalized policies evaluated for 100 episodes across 5 seeds reported in terms of mean +/- std over rewards for all environments except MO-Kitchen, in which task success rate is used as metric}. 
For this table only, we abbreviate Privileged Traj-MaskedDP to \textbf{Privileged} and  Anonymized CLS-MaskedDP to \textbf{Anonymized} consistent with the rest of the paper. 
% Both Privileged-RL and Anonymized-OnlineRL achieve stronger performance than the baseline methods, showing that P3L yields better-aligned personalized policies. 
Privileged-RL achieves the best performance across most of the environments, consistently demonstrating strong performance relative to the baselines,
% Privileged-RL achieves the best performance across all environments, consistently outperforming the baselines,
while Anonymized-RL remains competitive with or surpasses the baselines without access to annotator identity, suggesting that P3L generates policies that are better aligned with individual user preferences.
Blue denotes the best result and gray the second-best.}
 % Both our frameworks,  and CLS-MaskedDP achieve higher average RL performance than the baselines, demonstrating improved policy alignment.}
  \label{All-RL-table}
\end{table*}

% \textcolor{red}{REWRITE ALL----ASK ABOUT NAMING - USING ONLINE RL INSTEAD OF PPO/ARS AND SHOULD FRANKA KITCHEN BE A SEPARATE TABLE-HOW DO O WRITE IN TABLE ABOUT IT BEING OFFLINE OPTIMIZAITON ONLY- MENTIONED IN TEXT ALREADY UNDER BASELINES}
Table \ref{All-RL-table} compares performance of Online RL policies 
across all frameworks except for VPL, where we also optimized an offline IQL policy, for fair comparison to ~\citet{vpl_poddar2024personalizing} who used IQL.
We report the average RL performance as mean and standard deviation over reward obtained over 100 rollouts across 5 seeds across all users and all seeds per model for all environments except MO-Kitchen where we report mean and standard deviation over task success rate. 
For more results with respect to performance across individual new users, please refer to Appendix ~\ref{app:RL_exp_details}.
% Updateseeds if changed for kicthen and mention Task success rate and only IQL for FrankaKitchen and talk about Humanoid-finishing IQL on this first.}
We observe from Table ~\ref{All-RL-table} that across all environments, both of our frameworks, for the majority of users on average, deliver a more
personalized policy than both the baselines.

% For MO-Kitchen-2d and MO-Kitchen-4d, the Anonymized-CLS-MaskedDP outperforms the Single Reward Model but fails to improve over the VPL while having no access to annotator id,  while the Privileged-Traj-MaskedDP, which makes same assumption as VPL about complete access to priilige information about annotator id achieves substantially higher task success than all frmaeorosk.
% Interestingly, we see a drop in performance/task success rate across all frameworks as the complexity of the environment increases from MO-Kitchen-2d to MO-Kitchen-3d to MO-Kitchen-4d, which intutively aligns with our hypothesis that increasing no of objectives makes personalizaiton harder in emobied control. 

In MO-Highway (Row 1 in Table ~\ref{All-RL-table}), we observe that VPL obtains a higher score when learning with per-timestep offline RL IQL policy than any other models
which optimize an online RL episodic policy. 
However, this performance appears to stem from the offline IQL policy optimization rather than the VPL reward model itself since optimizing an online ARS policy under the VPL reward, fails to achieve comparable performance.
In all other environments, Privileged-Traj-MaskedDP, which operates under the same assumption as VPL of complete access to annotator identity, substantially outperforms VPL and Single Reward Model baselines.

Moreover, P3L remains effective when this privileged information is removed: although Anonymized-CLS-MaskedDP faces a harder learning problem and does not consistently outperform VPL, it remains competitive despite having no access to annotator identity and still surpasses the Single Reward Model baseline. 
Together, these results demonstrate that P3L benefits from privileged information when available while retaining strong performance under weaker and more realistic information assumptions.

As the environment becomes increasingly complex from MO-Kitchen-2d  to MO-Kitchen-3d to MO-Kitchen-4d (Row 4-6 respectively in Table ~\ref{All-RL-table}), task success declines across all evaluated frameworks. 
This systematic decrease in performance supports our hypothesis that learning personalized policies becomes increasingly difficult as the dimensionality of the underlying preference space grows.
Despite the growing complexity, the strong relative performance of P3L across these settings reflects its effectiveness for enabling personalized policies in embodied-control tasks.
% its effectiveness as the number of objectives increases.
% personalization becomes more challenging as the number of objectives increases in embodied control.
% Importantly, the advantage of P3L persists across these increasingly difficult settings, with Privileged-Traj-MaskedDP showing the strongest performance when methods are compared under matched assumptions.

% However, this performance appears to stem from the offline IQL policy optimization rather than the VPL reward model itself since optimizing an online RL policy under the exact same learned VPL reward, fails to achieve comparable performance. 

% However, this performance appears to stem from the offline IQL policy optimization rather than the VPL reward model itself since optimizing an online ARS policy under the VPL reward, fails to achieve comparable performance. 
% like our frameworks, does not deliver an equally high performance. 
% However, this improvement appears to stem from the offline policy rather than the VPL model itself, as VPL does not when used to optimize an online ARS policy.

\begin{figure}[t]
\centering

\begin{subfigure}{\linewidth}
\centering
\resizebox{\linewidth}{!}{
\begin{tabular}{lccc}
\toprule
 & \multicolumn{1}{c}{Baseline} & \multicolumn{2}{c}{P3L} \\
\cmidrule(lr){2-2} \cmidrule(lr){3-4}
Environment  & VPL & Privileged-Traj-MaskedDP & Anonymized-CLS-MaskedDP \\
\midrule
MO-Highway   &131.63 $\pm$ 3.19 &\best{0.75 $\pm$  0.49} &  \second{2.16 $\pm$ 0.20}   \\
\midrule
 MO-Reacher & 225.21 $\pm$ 9.33
& \second{7.83 $\pm$ 0.97}
& \best{4.16 $\pm$ 0.06} \\
\midrule 
MO-Humanoid  & 18018.25 $\pm$ 3784.13& \best{7.23 $\pm$ 0.88}   & \second{14.07 $\pm$ 3.55}\\
% \midrule
%  MO-Reacher & 225.21 $\pm$ 9.33
% & \second{7.83 $\pm$ 0.97}
% & \best{4.16 $\pm$ 0.06} \\
% \midrule 
% MO-Humanoid  & &    & \\
% \midrule
% Franka-Kitchen -DECIDE WRT SEEDS & &    & \\
\bottomrule
\end{tabular}
}
\caption{Training Time}
\label{fig:train-time-table}
\end{subfigure}

\vspace{0.5em}

\begin{subfigure}[t]{\linewidth}
\centering
\includegraphics[width=0.75\linewidth]{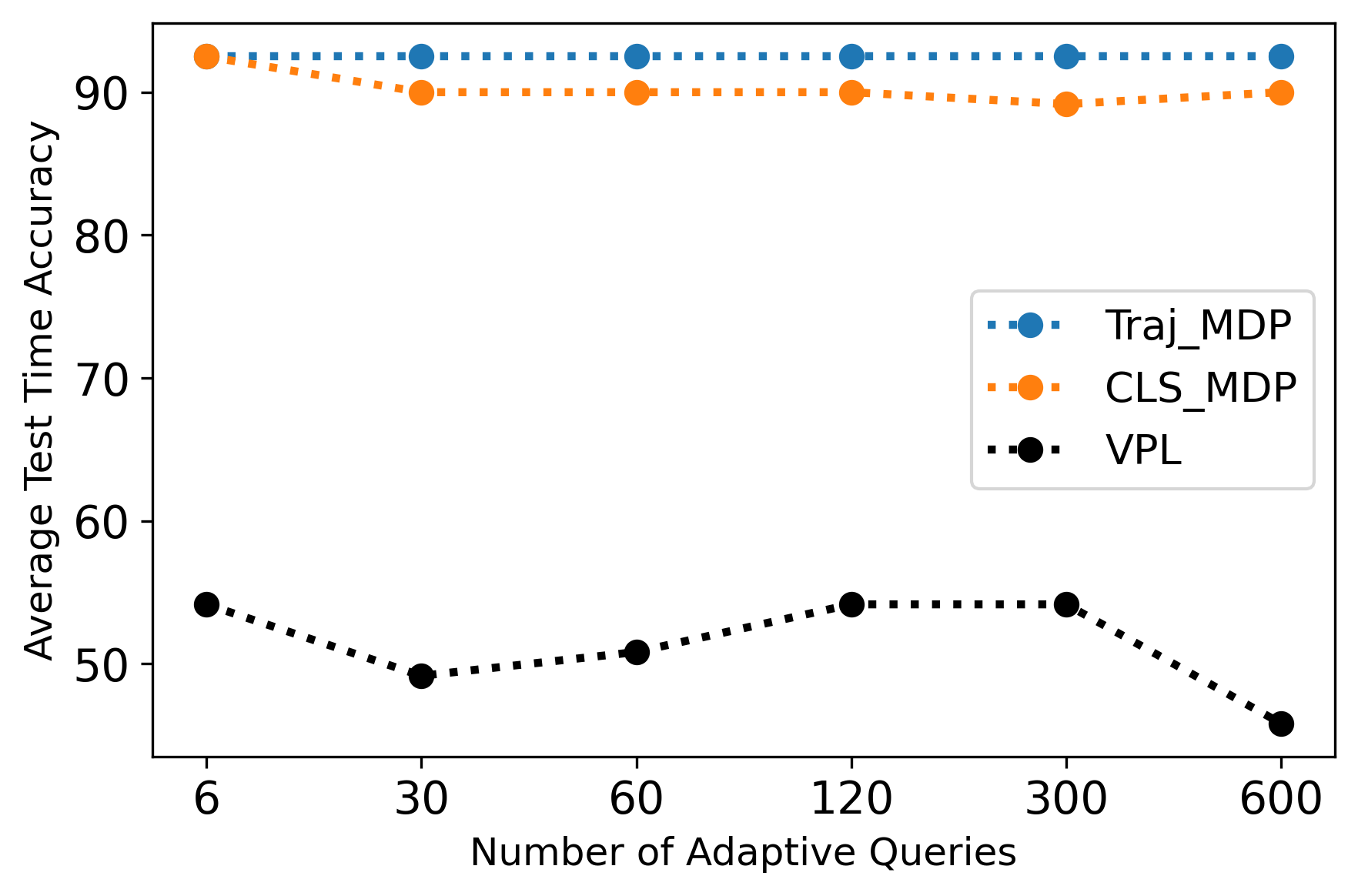}
\caption{Alignment vs Adaptation Query Budget}
\label{fig:adapt_ablation}
\end{subfigure}
% \caption{}
% \label{}
% \end{figure}
% \begin{figure*}[t]
% \centering
% % \footnotesize

% \begin{subfigure}[t]{0.51\linewidth}
% \centering
% \resizebox{\linewidth}{!}{
% \begin{tabular}{lccc}
% \toprule
%  & \multicolumn{1}{c}{Baselines} & \multicolumn{2}{c}{MaskedDP} \\
% \cmidrule(lr){2-2} \cmidrule(lr){3-4}
%  & VPL & Traj-MaskedDP & CLS-MaskedDP \\
% \midrule
% Training Time & 225.21 $\pm$ 9.33  & \second{7.83 $\pm$ 0.97} & \best{4.16 $\pm$ 0.06} \\
% \bottomrule
% \end{tabular}

% }
% % \vspace{1cm}
% \caption{}
% \label{fig:train-time-table}
% \end{subfigure}
% \hfill
% \begin{subfigure}[t]{0.45\linewidth}
% \centering
% % Placeholder for bar plot
% \includegraphics[width=\linewidth]{figures/New-adapt-ablation-avg_accuracy_queries.pdf}
% \caption{}
% \label{fig:adapt_ablation}
% \end{subfigure}

\caption{(a) \textbf{Training time} in seconds reported as mean and standard deviation across 5 seeds: the lower time it takes, the faster the framework is. (b) \textbf{Test-time accuracy vs. the number of queries used for adaptation} demonstrates that both our frameworks Traj-MaskedDP and CLS-MaskedDP achieve high accuracy around 90\% with smaller number of preference queries as with bigger adaptation query set size whereas VPL fails to achieve a comparable performance even when fed 100x adaptation queries per user.    }

\label{fig:train-time}

\end{figure}

\paragraph{Training Time Comparison}
We report the time taken to train the pluralistic frameworks in Fig.  ~\ref{fig:train-time-table} on Nvidia GeForce RTX 3090 GPU. 
From the table we see that VPL incurs the highest training cost whereas both P3L variants substantially outperform VPL in training efficiency across all three environments, achieving speedups ranging from approximately 29× to nearly 2,500×. Notably, Privileged-Traj-MaskedDP trains approximately 176× faster on MO-Highway, 29× faster on MO-Reacher, and nearly 2,500× faster on MO-Humanoid. Even without access to annotator identities, Anonymized-CLS-MaskedDP achieves speedups of 61×, 54×, and over 1,000×, respectively. 
%  taking up to 50× longer to train than Anonymized-CLS-MaskedDP and up to 30× longer than Privileged-Traj-MaskedDP, even though Privileged-Traj-MaskedDP makes similar assumptions. 
% Anonymized-CLS-MaskedDP is consistently the fastest, requiring the least training time as reported across 5 seeds.
% We see that VPL takes as much as 50 times the time taken to train  CLS-MaskedDP and as much as 30 times the time taken to train Traj-MaskedDP which makes similar assumptions to VPL . Our CLS approach is the fastest, requiring least time to train  
% Our non-privileged approach uses attention for final summarization and thus takes longer that our privileged approach , which is fastest across all frameworks. 
This has significant practical implications with respect to computational cost, energy consumption, and compute budget.

% \subsubsection{Adaptive queries ablation}
\paragraph{Test-Time Preference Alignment vs. Adaptation Query Budget}

In Fig ~\ref{fig:adapt_ablation}, we compare and contrast the impact of number of adaptive queries on test accuracy across our approaches against VPL on MO-Reacher.
% For both our frameworks, Traj-MaskedDP as well as CLS-MaskedDP, we can use any number of preference queries for adaption to new user whereas VPL requires a fixed number of queries  to infer the latent preference at test time for a new user. 
% 
Both of our frameworks, Privileged-Traj-MaskedDP and Anonymized-CLS-MaskedDP, adapt to a new user using any number of preference queries, allowing flexible adaptation at test time. 
By contrast, VPL requires a fixed number of queries to infer the latent preference for a new user. This stems from the VPL architecture which uses a VAE with a linear layer as first layer, rendering it unable to handle variable-sized adaptation query sets.
Consequently, we train a new VPL model for each adaptation query size. 
Our results demonstrate that both our frameworks, not only work with any variable size of adaptation queries but can also achieve competitive performance of 90\% accuracy with smaller number of adaptation queries as they obtain with bigger query sets;
whereas VPL remains highly sensitive to the adaptation size and performs substantially worse even with significantly larger query sets of 100x per user. For per-user adaptation ablation, please ref to Appendix~\ref{app:per_user_adapative_ablation}.
% Whereas VPL is very sensitive to adaption set size and despite increasing adaption set size to 100x per user, fails to produce high accuracy to test time user. 

\subsection{Personalization to Out-of-Distribution New Users}
We explore personalization to out-of-distribution (OOD) new users at test time by training on any possible subset of different users in MO-Reacher and MO-Highway, and then testing adaptation to a new user with an unseen behavior, both with respect to test set accuracy as well as the RL performance.  
For adaptation to a new user, to get $k_j$ for our frameworks and to infer latent preference variable for VPL, we use 30 adaptation queries.
% And since this is an online adaptation to new user out of training distribution, we only optimize an RL ARS policy under VPL unlike in-distribution VPL-IQL.
% Table~\ref{tab:OOD} summarizes OOD adaptation performance over 5 seeds, using 15 test preference queries and 100 RL rollouts with only 30 adaptation queries.

We report mean and standard deviation across 5 seeds over accuracy on test preference queries and 100 rollouts for RL for the two metrics respectively in Table ~\ref{tab:OOD_avg}. For more details, please see appendix ~\ref{app:more_ood_eval}.
Our approach, Privileged-Traj-MaskedDP outperforms VPL on both test set accuracy as well as on downstream RL performance under the same annotator-identity assumption. 
It also outperforms Anonymized-CLS-MaskedDP, which makes no such assumption and is therefore trained under weaker conditions.
Notably, despite being trained without annotator identity information, Anonymized-CLS-MaskedDP still achieves higher test as well as RL performance than VPL when adapting to new OOD user. 

\begin{table*}[t!]
\centering
% \footnotesize
\scriptsize
% \tiny
\setlength{\tabcolsep}{4 pt}
% \centering
% \resizebox{\linewidth}{!}{
\begin{tabular}{llrrr}
\toprule
Environment &
Eval Metric & \multicolumn{1}{c}{Baseline} &
\multicolumn{2}{c}{P3L} \\
\cmidrule(lr){3-3}
\cmidrule(lr){4-5}

& & \multicolumn{1}{c}{VPL}
& \multicolumn{1}{c}{Privileged-Traj-MaskedDP}
& \multicolumn{1}{c}{Anonymized-CLS-MaskedDP} \\
\midrule 

\multirow{2}{*}{MO-Highway} &
Test
&  42.667 ± 30.346 
& \best{81.333 ± 9.978}
& \second{78.000 ± 8.641} \\

                       & RL & -5.270 ± 12.064 & \second{4.866 ± 20.605}  & \best{ 7.119 ± 17.783} \\
                      
\midrule
\multirow{2}{*}{MO-Reacher} &
Test
&  42.333 ± 36.189 
& \best{$100.00 \pm 0.00$}
& \second{$94.333 \pm 9.815$} \\
% \midrule
& RL
&  19.598 ± 2.251
& \best{$32.552 \pm 1.585$}
& \second{$22.717 \pm 5.334$} \\
% \midrule
% \multirow{2}{*}{MO-Humanoid}& Test &  71.900 ± 1.281 
% & \best{ 94.600 ± 1.319}
% & \second{94.600 ± 1.393} \\

%                       &  RL & \best{17.394 ± 184.277} & \second{-30.597 ± 144.278}  &  -31.988 ± 160.191\\ 
                      
\bottomrule

\end{tabular}

% \bottomrule
% \end{tabular}
\caption{\textbf{Out-of-distribution adaptation to new users reported as mean ± std over 5 seeds for test preference accuracy and RL performance (100 evaluation rollouts)}. P3L generalizes better to unseen users, achieving higher preference accuracy and stronger downstream RL performance than VPL, both with and without access to annotator identity. }
\label{tab:OOD_avg}

\end{table*}

\section{Discussion and Conclusion}

We present a framework for pluralistic alignment that enables personalized policy learning from heterogeneous crowd feedback, both with and without privileged information about annotator identities. 
Experiments in multi-objective control environments show improved preference alignment, competitive RL performance, and significantly reduced training cost compared to prior approaches. These results suggest that learning structured representations of diverse human preferences provides a scalable path toward personalized alignment in RLHF systems. Unlike prior work ~\cite{vpl_poddar2024personalizing}, which assumes access to annotator IDs, our approach can operate effectively without this information while still achieving competitive performance in terms of both preference test-set accuracy and downstream RL performance.
A key advantage of our framework is its computational efficiency. Compared to existing approaches, our method significantly reduces training cost while maintaining strong alignment performance. This efficiency is particularly important in RLHF settings, where learning from human feedback is often sample-intensive~\citep{christiano2017deep,ouyang2022training}. P3L achieves high test-time accuracy while requiring substantially less training time.
Our method also supports flexible adaptation at test time thus lowering the human interaction burden at test time.

% Our method also supports flexible adaptation at test time. Because personalization relies on reweighting shared preference directions rather than retraining a latent user model, the framework can adapt using any number of preference queries. This flexibility removes the requirement that the same annotator remain available during test-time adaptation and allows efficient personalization with only a small number of queries, substantially reducing human interaction costs.

We demonstrate that P3L successfully aligns policies to users both in-distribution and out-of-distribution relative to the offline preference dataset. Notably, Anonymized-CLS-MaskedDP variant operates under the weaker assumption that annotator identities are unavailable, yet often outperforms VPL which relies on annotator identity information—across both RL performance and preference prediction accuracy.
More broadly, our results suggest that heterogeneous crowd feedback can be effectively leveraged by learning a shared representation of preference directions. Once this representation is learned, personalized reward functions for new users can be constructed efficiently from only a small number of adaptive queries, providing a scalable mechanism for pluralistic alignment in RLHF systems.

One limitation of our approach is pre-specification of the number of PCA components used to represent preference directions. In practice, we determine this value through ablation studies to capture the structure of the  underlying preference distribution. 
Future work can explore metwhods for automatically inferring number of latent objectives present in crowdsourced feedback in an unsupervised manner and also investigate incorporating richer feedback modalities, including demonstrations, rankings, and corrective feedback.
% Future work could explore methods for automatically inferring the number of latent objectives present in crowdsourced feedback in an unsupervised manner, removing the need to manually select the number of eigenvectors. 
% Another promising direction is extending P3L
% to incorporate additional feedback modalities such as demonstrations, rankings, or corrective feedback.

\section*{Acknowledgement}
This work has taken place in the Aligned, Robust, and Interactive Autonomy (ARIA) Lab at the University of Utah. ARIA Lab research is supported in part by the National Science Foundation (OIA-2535195, IIS2310759, IIS2416761), U.S. Army DEVCOM Army Research Laboratory (W911NF-19-S-0001), the NIH (R21EB035378), ARPA-H, and Coefficient Giving. Any opinions, findings, and conclusions expressed in this material are those of the authors and do not necessarily reflect the views of the sponsors.

\bibliography{v2_refs,aaai2027}
\setcounter{secnumdepth}{2}
\appendix
% \section{Appendix}
% \newpage
\clearpage
\section{Experimental Setup Details}~\label{app:preference_setup_details}

We use the Fast Highway variant with a fixed horizon of 30 timesteps with 3 lanes and 20 vehicles. 
For MO-Highway we used 300 pairwise training preference queries and 30 adaptive queries with number of PCA components set to 3 for both variants of P3L while in MO-Reacher we used only 60 adaptive queries and 600 pairwise training preference queries with number of PCA components set to 2 across both our frameworks.
For MO-Humanoid we used 1000 pairwise training queries and 250 adaptive queries with PCA dimension of 10.
For MO-Kitchen-2d we used 100 pairwise training queries and 15 adaptive queries with PCA dim of 2 while for MO-Kitchen-3d we used 900 pairwise training queries and 90 adaptive queries with PCA dim of 3 and lastly, for MO-Kitchen-4d we used 2400 pairwise training queries and 180 adaptive queries with PCA dim of 4.

\emph{For appendix only, please note that we abbreviate Privileged Traj-MaskedDP to \textbf{Traj-MaskedDP} and  Anonymized CLS-MaskedDP to \textbf{CLS-MaskedDP} consistent with the rest of the paper. }

\begin{table*}
\centering
\footnotesize
\setlength{\tabcolsep}{9pt}
 
  \centering
  \begin{tabular}{lrrrr}
 & \multicolumn{2}{c}{Baselines} & \multicolumn{2}{c}{MaskedDP} \\
 \cmidrule(lr){2-3} \cmidrule(lr){4-5}
    & 
    SingleRM    & VPL   & Traj-MaskedDP  & CLS-MaskedDP \\
    \midrule
User 1 &  68.000 ± 13.597  & 73.333 ± 22.311 & \best{92.000 ± 7.775} & \second{85.333 ± 11.470}\\
User 2 &  77.333 ± 9.97 & 77.333 ± 9.978 & \best{78.667 ± 13.597} & \second{77.333 ± 9.043} \\
User 3 & 65.333 ± 13.597 & 49.333 ± 26.866 & \best{84.000 ± 5.333}  & \second{73.333 ± 4.216} \\
\midrule
Average User Accuracy & 70.222 ± 3.326 & 66.667 ± 3.143 & \best{84.889 ± 5.145} &\second{78.667 ± 4.576}  \\
    \bottomrule
  \end{tabular} 
   \caption{\textbf{Accuracy of Held out Test-Set Preference queries on MO-Highway} across reported as mean and standard deviation across 5 seeds with 30 adaptive preference queries and 15 held out preference test queries. Both of our approaches, \textbf{TrajMaskedDP} which assumes access to privileged annotator ids (as does VPL), as well as \textbf{CLSMaskedDP} which does not assume any access to annotator identity, outperform VPL, for individual users, as well as on average. Single Reward Model achieves competitive performance for one user as our approaches but on average fails to align with each user. } \label{Highway-Acc-table}
\end{table*}

\begin{table*}[h]
\centering
\footnotesize
\setlength{\tabcolsep}{9pt}
  
  \centering
  \begin{tabular}{lrrrr}
 & \multicolumn{2}{c}{Baselines} & \multicolumn{2}{c}{MaskedDP} \\
 \cmidrule(r){2-3} \cmidrule(lr){4-5}
    & 
    SingleRM    & VPL  & Traj-MaskedDP  & CLS-MaskedDP \\
    \midrule
User 1 & 44.000 ± 2.494 & 27.333 ± 21.848 & \best{96.000 ± 6.464} & \second{87.333 ± 3.887} \\
User 2 & 65.333 ± 4.522 &  71.333 ± 18.086 &\best{93.333 ± 2.981} & \second{92.667 ± 2.494} \\
User 3 & 24.000 ± 3.887 & 52.667 ± 32.069 & \second{87.333 ± 6.464} & \best{99.333 ± 1.333} \\
User 4 & 85.333 ± 4.989 & 56.667 ± 35.777 &\best{95.333 ± 3.399} & \second{94.667 ± 2.667}  \\
\midrule
Average User Accuracy & 54.667 ± 0.667 & 52.000 ± 3.055 & \second{93.000 ± 1.453} & \best{93.500 ± 1.434}\\
    \bottomrule
  \end{tabular} 
    \caption{\textbf{ Accuracy of Held out Test-Set Preference queries on MO-Reacher}  across 5 seeds with 60 adaptive preference queries and 30 held out preference test queries. Both of our approaches,\textbf{Traj-MaskedDP} which assumes access to privileged annotator ids (as does  VPL), as well as \textbf{CLS-MaskedDP} which does not assume any access to annotator identity, outperform both the  baselines Single Reward Model and VPL, for individual users, as well as on average. The blue background indicates the best performance, while the gray background denotes the second-best performance.
 }\label{Reacher-Acc-table}

\end{table*}
\begin{table*}[h!]
\centering
\footnotesize
\setlength{\tabcolsep}{9pt}
  
  \centering
  \begin{tabular}{lrrrr}
 & \multicolumn{2}{c}{Baselines} & \multicolumn{2}{c}{MaskedDP} \\
 \cmidrule(r){2-3} \cmidrule(lr){4-5}
    & 
    SingleRM    & VPL  & Traj-MaskedDP  & CLS-MaskedDP \\
    \midrule
User 1 & 81.000 ± 5.099 & 76.600 ± 7.864 & \best{90.400 ± 2.939 } & \second{81.000 ± 2.191} \\
User 2 & 72.400 ± 5.678  & 74.800 ± 5.741 &\best{85.800 ± 3.059} & \second{82.400 ± 0.800} \\
User 3 & \second{73.200 ± 5.706} & 68.800 ± 6.046 & \best{88.000 ± 1.789} & 70.000 ± 2.191\\
User 4 & 71.400 ± 4.800 & 74.600 ± 6.343 &\best{89.000 ± 3.033} & \second{75.400 ± 2.653}  \\
User 5 & \second{83.400 ± 7.940} & 79.800 ± 6.524 &\best{88.400 ± 3.499} & 80.600 ± 2.653 \\
\midrule
Average User Accuracy & 76.280 ± 1.294 & 74.920 ± 2.018 & \best{88.320 ± 0.515} & \second{77.880 ± 1.017}\\
    \bottomrule
  \end{tabular} 
    \caption{\textbf{ Accuracy of Held out Test-Set Preference queries on MO-Humanoid}  across 5 seeds with 250 adaptive preference queries and 100 held out preference test queries. Both of our approaches,\textbf{Traj-MaskedDP} which assumes access to privileged annotator ids (as does VPL), as well as \textbf{CLS-MaskedDP} which does not assume any access to annotator identity, outperform both the  baselines Single Reward Model and VPL, for most of individual users, as well as on average . The blue background indicates the best performance, while the gray background denotes the second-best performance.
 }\label{Humanoid-Acc-table}

\end{table*}

\section{Detailed Per-User Test Accuracy }\label{app:all-env-per-user-test-acc}
\subsection{MO-Highway}\label{app:Indv-Highway-Test-Acc}
For each user with a different preference objective, we report the results in Table ~\ref{Highway-Acc-table} across 5 seeds for each user with a different preference objective. P3L outperforms the baselines across individual users as well as on average.

\subsection{MO-Reacher} \label{app:Indv-Reacher-Test-Acc}
In Table~\ref{Reacher-Acc-table}  for MO-Reacher, we present the accuracy of adapting to test-time held out preference queries in terms of mean and standard deviation across 5 seeds for each user with a different preference objective. Both of our P3L approaches, Privileged-Traj-MaskedDP which assumes privilege annotator id
is given like VPL, as well as Anonymized-CLS-MaskedDP which does not assume any access to annotator identity, outperforms both the
baselines Single Reward Model and VPL, for individual users, as well as on average.
% \textcolor{red}{To be updated-For MO-humanoid Table~\ref{Humanoid-Acc-table} and for Franka Kitchen~\ref{Kitchen-Acc-table} respectively represent accuracy of adapating to held out test preferences.- Will update results lines below once I have accuracy ready across all seeds.}

\subsection{MO-Humanoid}\label{app:Indv-humanoid-test-acc}
For each user with a different preference objective , we report the accuracy of Held out Test-Set Preference queries on MO-Humanoid across 5 seeds in Table ~\ref{Humanoid-Acc-table} . We observe that for our approaches Privileged-Traj-MaskedDP and Anonymized-CLS-MaskedDP outperform VPL, where Privileged-Traj-MaskedDP makes same assumptions about prior knowledge of annotator identity as VPL while Anonymized-CLS-MaskedDP approach has no such prior knowledge. 

\subsection{MO-Kitchen}\label{app:Indv-FrankaKitchen-Test-Acc}
We report held-out pairwise preference test accuracy across 5 seeds for each user in increasingly complex MO-Kitchen variants: MO-Kitchen-2d, involving two users (Microwave and Kettle), with 15 adaptive and 25 held-out queries in Table~\ref{Kitchen-Acc-table-2d}; MO-Kitchen-3d, involving three users (Microwave, Kettle, and Light Switch), with 90 adaptive and 45 held-out queries in Table~\ref{Kitchen-Acc-table-3d}; and MO-Kitchen-4d, involving four users (Microwave, Kettle, Light Switch, and Top Burner), with 180 adaptive and 90 held-out queries in Table~\ref{Kitchen-Acc-table-4d}.

Across all three MO-Kitchen variants, both P3L approaches consistently outperform the baselines in held-out pairwise preference prediction accuracy. Privileged-Traj-MaskedDP achieves the highest average accuracies of 100.0\%, 84.148\%, and 78.444\% on MO-Kitchen-2d, MO-Kitchen-3d, and MO-Kitchen-4d, respectively, while CLS-MaskedDP achieves 90.4\%, 82.667\%, and 72.167\%. 
Notably CLS-MaskedDP, which does not assume access to annotator identities, also substantially outperforms VPL, despite VPL having access to such privileged information, highlighting the effectiveness of our framework in learning user-specific preferences from anonymized feedback.
Although the performance of both approaches decreases as the number of users increases, they consistently maintain substantially higher accuracies than Single Reward Model and VPL, whose performance remains close to 50–55\% across the three settings.

% CLS-MaskedDP achieves strong performance without requiring access to annotator identities, highlighting its effectiveness in learning user-specific preferences from anonymized feedback.
%  Notably, CLS-MaskedDP maintains this advantage despite not requiring annotator identity information, whereas VPL assumes access to such information.
% Although accuracy decreases for both approaches as the number of users and task complexity increase, they maintain substantial performance advantages over the baselines. Notably, CLS-MaskedDP achieves strong performance without requiring access to annotator identities, 

% We observe that P3L, under both Privileged and Anonymized setting shows better alignment than both the baselines. 

\begin{table*}[t]
\centering
\footnotesize
\setlength{\tabcolsep}{9pt}
  
  \centering
  \begin{tabular}{lrrrr}
 & \multicolumn{2}{c}{Baselines} & \multicolumn{2}{c}{MaskedDP} \\
 \cmidrule(r){2-3} \cmidrule(lr){4-5}
    & 
    SingleRM    & VPL  & Traj-MaskedDP  & CLS-MaskedDP \\
    \midrule
User 1  &48.800 ± 6.400 & 63.200 ± 39.387  & \best{100.000 ± 0.000} & \second{90.400 ± 4.079}\\
User 2  & 51.200 ± 6.400 & 37.600 ± 38.831 & \best{ 100.000 ± 0.000}  & \second{90.400 ± 4.079}\\

\midrule
Average User Accuracy & 50.000 ± 0.000
 &  50.400 ± 0.800 & \best{100.000 ± 0.0 } & \second{90.400 ± 4.079}\\
    \bottomrule
  \end{tabular} 
    \caption{\textbf{ Accuracy of Held out Test-Set Preference queries on MO-Kitchen-2d}  across 5 seeds with 15 adaptive preference queries and 25 held out preference test queries. Both of our approaches,\textbf{Traj-MaskedDP} which assumes access to privileged annotator ids (as does VPL), as well as \textbf{CLS-MaskedDP} which does not assume any access to annotator identity, outperform both the  baselines Single Reward Model and VPL, for most of individual users, as well as on average. The blue background indicates the best performance, while the gray background denotes the second-best performance.
 }\label{Kitchen-Acc-table-2d}

\end{table*}

\begin{table*}[t]
\centering
\footnotesize
\setlength{\tabcolsep}{9pt}
  
  \centering
  \begin{tabular}{lrrrr}
 & \multicolumn{2}{c}{Baselines} & \multicolumn{2}{c}{MaskedDP} \\
 \cmidrule(r){2-3} \cmidrule(lr){4-5}
    & 
    SingleRM    & VPL  & Traj-MaskedDP  & CLS-MaskedDP \\
    \midrule
User 1  &  72.000 ± 12.696& 47.111 ± 16.354  & \best{ 84.444 ± 4.869} & \second{84.444 ± 3.718}\\
User 2  &  36.889 ± 12.927  &  37.778 ± 9.428 & \best{82.667 ± 2.948}  & \second{79.556 ± 7.221}\\
User 3  & 50.667 ± 6.029  & 69.333 ± 17.855 & \best{ 85.333 ± 3.875}  & \second{84.000 ± 3.266}\\

\midrule
Average User Accuracy & 53.185 ± 1.582 &  51.407 ± 1.525 & \best{84.148 ± 2.227 } & \second{82.667 ± 3.647}\\
    \bottomrule
  \end{tabular} 
    \caption{\textbf{ Accuracy of Held out Test-Set Preference queries on MO-Kitchen-3d}  across 5 seeds with 90 adaptive preference queries and 45 held out preference test queries. Both of our approaches,\textbf{Traj-MaskedDP} which assumes access to privileged annotator ids (as does VPL), as well as \textbf{CLS-MaskedDP} which does not assume any access to annotator identity, outperform both the  baselines Single Reward Model and VPL, for most of individual users, as well as on average. The blue background indicates the best performance, while the gray background denotes the second-best performance.
 }\label{Kitchen-Acc-table-3d}
\end{table*}

\begin{table*}[h!]
\centering
\footnotesize
\setlength{\tabcolsep}{9pt}
  
  \centering
  \begin{tabular}{lrrrr}
 & \multicolumn{2}{c}{Baselines} & \multicolumn{2}{c}{MaskedDP} \\
 \cmidrule(r){2-3} \cmidrule(lr){4-5}
    & 
    SingleRM    & VPL  & Traj-MaskedDP  & CLS-MaskedDP \\
    \midrule
User 1  & 65.778 ± 10.242& 54.667 ± 17.104& \best{79.111 ± 4.467}& \second{75.556 ± 4.332}\\
User 2 &  51.333 ± 8.025& 48.444 ± 10.390& \best{73.778 ± 3.823}& \second{69.333 ± 5.287}\\
User 3  & 42.889 ± 5.287& 40.222 ± 9.671& \best{77.556 ± 2.931}& \second{67.333 ± 3.414}\\
User 4  & 61.111 ± 10.352 & 63.556 ± 4.838 & \best{83.333 ± 2.534} & \second{76.444 ± 3.326} \\
\midrule
Average User Accuracy & 55.278 ± 2.159& 51.722 ± 2.905 &\best{78.444 ± 0.572} & \second{72.167 ± 2.021}\\
    \bottomrule
  \end{tabular} 
    \caption{\textbf{ Accuracy of Held out Test-Set Preference queries on MO-Kitchen-4d}  across 5 seeds with 180 adaptive preference queries and 90 held out preference test queries. Both of our approaches,\textbf{Traj-MaskedDP} which assumes access to privileged annotator ids (as does VPL), as well as \textbf{CLS-MaskedDP} which does not assume any access to annotator identity, outperform both the  baselines Single Reward Model and VPL, for most of individual users, as well as on average. The blue background indicates the best performance, while the gray background denotes the second-best performance.
 }\label{Kitchen-Acc-table-4d}
\end{table*}

\section{Ablation on Annotation Set Size for VPL}
For VPL, we tried to tune the annotation set size, also referred to as context set size in ~\cite{vpl_poddar2024personalizing} for each domain.
% For MO-Reacher we swept over values of 6,30,120,300 and 600 to feed the chosen value no of queries at once to infer the best possible latent user preference $z$ from VAE, 
For MO-Reacher, we swept over annotation set sizes of 6, 30, 120, 300, and 600 preference queries, where the selected number of queries are provided at once to the VAE to infer the latent user preference vector $z$ which is then used to test alignment to new user both using test set accuracy as well as for optimizing an RL policy. 

For MO-Humanoid, we tuned the learning VPL over annotation set sizes of 1,5,25 and using an annealer. We finally chose to use set size of 1 as we observed posterior collapse with bigger annotation set sizes. 
For Kitchen, we started with annotation set size of 30 and soon observed posterior collapse so reduced annotation set size to 20 to successfully learn the VPL-VAE encoder and decoder reward. 

\section{In-distribution RL experiment details}\label{app:RL_exp_details}
We optimize RL policies under learned reward function using Augmented Random Search ~\citep{ars} and Proximal Policy Optimization(PPO) ~\citep{schulman2017proximal} from Stable Baselines3 ~\citep{sb3} with episodic rewards for 7e6 timesteps for MO-Reacher and with 2e5 timesteps for MO-Highway respectively. Note, we use Stable-Baselines3 Contrib for ARS. 
Original VPL is an offline framework which uses Implicit Q-Learning (IQL) \citep{kostrikov2022offline} to learn an offline policy optimized for per-timestep reward. 
From TorchRL~\citep{bou2024torchrl}, we use discrete IQL for MO-Highway and MO-Reacher and continuous IQL for MO-Humanoid and MO-Kitchen respectively.

MO-Reacher is set to have a fixed horizon of 50 and does not have any terminal condition by default. We use ARS with default parameters in SB3 to optimize an RL policy using episodic rewards.
MO-FastHighway on the other hand by default ends if there's a collision or if lapses 30 timesteps. So to avoid early termination bias, we ignore the terminal condition and set the environment to have a 
fixed horizon of 30. 
We use Vectorized Norm with $norm\_obs=True$ and $norm\_reward=True$ from SB3 to train PPO policies with normalized obs and normalized rewards in Fast Highway. 
% ADD DETAILS ABOUT MO-HUMANOID, IQL DETIALS AND FRANKA-KITCHEN
For MO-Humanoid , we optimized ARS policy under learned reward for 10M timesteps using an MLP with learning rate  $0.02$ with n\_delta=16 and n\_top=8  and delta\_std=0.025.
For MO-Kitchen we optimize a Continuous IQL policy for 1M total timesteps with learning rate =0.0003 , expectile = 0.7, temperature=3.0 and batch size of 256. 

Table \ref{All-RL-table} compares average performance of RL policies  learned using PPO as online RL algorithm for MO-Highway and  ARS as online RL algorithm for MO-Reacher and MO-Humanoid across all frameworks except for VPL, where we also optimized an offline IQL policy, for fair comparison to ~\citet{vpl_poddar2024personalizing} who used IQL; 
while the following subsections report for each environment, the RL performance across individual new users with different underlying preferences.

\subsection{MO-Highway In-distribution RL}
From Table \ref{Highway-RL-table}, we observe for 2 users (User 1 and User 2) who prefer high speed driving and driving in right most lane respectively, that RL performance under CLS-MaskedDP is highest. 
This implies that CLS-MaskedDP yields a more personalized policy than any other model, even though it has no prior information about annotator identity unlike VPL and Traj-MaskedDP. 
User 3 prefers driving most conservatively to avoid collision, and interestingly we notice that Single Reward Model here does better than any other personalized framework. 
We hypothesize this might be due to the limited diversity of trajectories in the offline preference dataset for this behavior, since this behavior uses sparse rewards: 
$0$ for avoiding collisions and $-1$ for collisions.
Additionally, since the environment is fixed horizon and does not terminate on collision, negative rewards for this user keep accumulating. 
But on average across all users, RL performance of our privileged approach Traj-MaskedDP, still outperforms all baselines. 

\begin{table*}
\centering
\footnotesize
\setlength{\tabcolsep}{1pt}
  \caption{\textbf{RL performance MO-Highway} presented in terms of mean and standard deviation across 5 seeds with 100 rollouts per seed. We achieve competitive performance as compared to baselines on individual users and on average except User 3 which prefers strictly conservative driving. }
  \label{Highway-RL-table}
  \centering
  \begin{tabular}{lrrrrr}
  % \toprule
 & \multicolumn{3}{c}{Baselines} & \multicolumn{2}{c}{MaskedDP} \\
 \cmidrule(lr){2-4} \cmidrule(lr){5-6}
    & 
    SingleRM-PPO    & VPL-OfflineRL-IQL   &VPL-OnlineRL-PPO   & Traj-MaskedDP-PPO  & CLS-MaskedDP-PPO \\
    \midrule
 User 1 & 3.335 ± 4.459 & 2.247 ± 3.455 & 4.828 ± 4.125  & \second{5.120 ± 3.651} & \best{6.105 ± 3.984}\\
User 2 &  15.850 ± 12.777 &  \second{29.150 ± 3.603}  &  26.359 ± 8.281 & 29.024 ± 3.604 & \best{29.736 ± 1.043} \\
User 3 & \second{-6.276 ± 9.282} & \best{-6.170 ± 10.090} & -18.126 ± 10.030 & -10.980 ± 10.782 & -20.462 ± 7.263 \\
\midrule
Average RL performance &   4.303 ± 9.059 & \best{8.409 ± 15.063} & 4.354 ± 18.164 &  \second{7.721 ±  16.435} &  5.126 ± 20.505 \\

    \bottomrule
  \end{tabular} 
  
\end{table*}
 
\subsection{MO-Reacher In-distribution RL}

\begin{table*}[t!]
\centering
\footnotesize
\setlength{\tabcolsep}{1pt}
  \caption{\textbf{RL Performance on MO-Reacher} when evaluated over 100 episodes across 5 seeds reported in terms of mean +/- std. Both our frameworks, Traj-MaskedDP and CLS-MaskedDP achieve higher average RL performance than the baselines, demonstrating improved policy alignment.}
  \label{Reacher-RL-table}
  \centering
  \begin{tabular}{lrrrrr}
 & \multicolumn{3}{c}{Baselines} & \multicolumn{2}{c}{MaskedDP} \\
 \cmidrule(lr){2-4} \cmidrule(lr){5-6}
    & 
    SingleRM-ARS    & VPL-OfflineRL-IQL   & VPL-OnlineRL-ARS  & Traj-MaskedDP-ARS  & CLS-MaskedDP-ARS \\
    \midrule
User 1 & 13.871 ± 7.598 & 1.491 ± 7.190 &  19.721 ± 2.142 & \best{37.845 ± 6.244} & \second{29.024 ± 7.352}  \\
User 2 & 19.369 ± 8.004 &   \best{35.682 ± 13.034}& 21.106 ± 1.262 &\second{31.247 ± 4.934} & 31.216 ± 1.413 \\
User 3 & 17.436 ± 6.164 & 12.407 ± 5.575 & 22.618 ± 0.453
& \best{31.031 ± 3.827} & \second{26.740 ± 6.324}
 \\
User 4 & 16.140 ± 6.729 &  14.780 ± 12.437 &  18.262 ± 1.990 & \second{26.708 ± 1.444} & \best{27.432 ± 4.787}
 \\
 \midrule
Average & 16.704 ± 1.999 & 16.090 ± 12.372 & 20.427 ± 1.616 & \best{31.708 ± 4.520} & \second{28.603 ± 1.996} \\
    \bottomrule
  \end{tabular} 
  
\end{table*}
We observe from Table~\ref{Reacher-RL-table} that for MO-Reacher, both of our frameworks, for most users on average, deliver a more personalized policy than VPL \space or\space Single Reward Model except for User 2, where  VPL gets a higher score when learning with per-timestep offline RL IQL policy than any other models which optimize an online RL episodic policy.
However, this performance appears to stem from the offline IQL policy optimization rather than the VPL reward model itself since optimizing an online ARS policy under the VPL reward, fails to achieve comparable performance.

\subsection{MO-Humanoid In-distribution RL}
From Table~\ref{Humanoid-RL-table}, we can observe that P3L performs competitively across both settings with respect to the presence of annotator ID than baselines on individual new users as well as on average.
\begin{table*}[t!]
\centering
\footnotesize
\setlength{\tabcolsep}{1pt}
  \caption{\textbf{RL Performance on MO-Humanoid} when evaluated over 100 episodes across 5 seeds reported in terms of mean +/- std. Our framework outperforms all baselines across majority of users and on average too.}
  % \textcolor{red}{Tentative caption-Both our frameworks, Traj-MaskedDP and CLS-MaskedDP achieve higher average RL performance than the baselines, demonstrating improved policy alignment. --->Going to see how SingleRM on user 1 does for 5 or 10 seeds}}
  \label{Humanoid-RL-table}
  \centering
  \begin{tabular}{lrrrrr}
 & \multicolumn{3}{c}{Baselines} & \multicolumn{2}{c}{MaskedDP} \\
 \cmidrule(lr){2-4} \cmidrule(lr){5-6}
    & 
    SingleRM-ARS    & VPL-OfflineRL-IQL   & VPL-OnlineRL-ARS  & Traj-MaskedDP-ARS  & CLS-MaskedDP-ARS \\
    \midrule
User 1 &  28.995 ± 14.759 & \best{46.603 ± 49.040} & 8.323 ± 20.652 & \second{44.243 ± 31.841} &21.670 ± 12.302\\
User 2 &  5.452 ± 30.842 &  -9.016 ± 27.408 &20.551 ± 36.566  &  \best{74.810 ± 70.151}  &  \second{21.670 ± 12.302}\\
User 3 & 55.512 ± 21.987 &  \second{65.261 ± 47.616} &  63.180 ± 16.044&  \best{93.514 ± 57.554} & 60.124 ± 18.740 \\
User 4 &56.822 ± 26.341 &  30.802 ± 9.216 &  64.996 ± 44.143 &  \second{82.556 ± 44.696}  & \best{ 85.607 ± 12.131}  \\
User 5 & -388.945 ± 6.111 &  -390.156 ± 8.367 &-388.251 ± 8.060  & \best{-347.668 ± 42.936} &  \second{ -383.115 ± 4.540} \\
 \midrule
Average &  -48.433 ± 171.307  &  -51.301 ± 171.186 & -46.240 ± 172.485 & \best{-10.509 ± 169.372}& \second{-38.794 ± 173.861} \\
    \bottomrule
  \end{tabular} 
  
\end{table*}

\subsection{MO-Kitchen In-distribution RL}
We report RL performance in terms of task success rate for each user over 100 episodes across 5 seeds for each MO-Kitchen variant: MO-Kitchen-2d, consisting of two users (Microwave and Kettle), in Table~\ref{Kitchen-RL-table-2d}; MO-Kitchen-3d, consisting of three users (Microwave, Kettle, and Light Switch), in Table~\ref{Kitchen-RL-table-3d}; and MO-Kitchen-4d, consisting of four users (Microwave, Kettle, Light Switch, and Top Burner), in Table~\ref{Kitchen-RL-table-4d}. 
P3L consistently demonstrates superior performance compared to the baselines across all three MO-Kitchen variants.
We observe that our framework, specifically the Privileged-Traj-MaskedDP approach  achieves the highest average task success rate, obtaining 0.834, 0.799, and 0.661 on MO-Kitchen-2d, MO-Kitchen-3d, and MO-Kitchen-4d, respectively.
It also achieves the highest success rate on nearly all individual user tasks, with the exception of the Microwave task in MO-Kitchen-3d, where CLS-MaskedDP performs best. 
We also observe that increasing the number of users introduces greater task complexity and reduces performance across all methods.
% , P3L consistently maintains considerably higher success rates, with the performance gap remaining substantial even in the most challenging four-user setting.
% We report RL performance across for each user across 5 seeds and 100 episodes for each MO-Kitchen variant with- MO-Kitchen-2d in Table~\ref{Kitchen-RL-table-2d} ,MO-Kitchen-3d in Table~\ref{Kitchen-RL-table-3d} and MO-Kitchen-4d in Table~\ref{Kitchen-RL-table-4d} respectively. 
MO-Kitchen-4d is the most challenging task with long-horizon manipulation, as evidenced by most frameworks failing to optimize a successful RL policy under learned reward in Table~\ref{Kitchen-RL-table-4d}. 
Only Privileged-Traj-MaskedDP learns a policy that achieves substantial success across all four user tasks,
% Only Privileged-Traj-MaskedDP is able to actually learn a policy that can successfully complete the desired tasks, 
while VPL, making the same assumptions about the presence of annotator identities, fails for 3 out of 4 users. The anonymized variant of P3L struggles too, and as expected, the single reward model struggles most.

\begin{table*}[t!]
\centering
\footnotesize
\setlength{\tabcolsep}{1pt}
  \caption{\textbf{RL Performance on MO-Kitchen-2d} when evaluated over 100 episodes across 5 seeds reported in terms of mean +/- std for Task Success Rate. Since Franka-Kitchen is an offline environment, we run only offline RL IQL.}
  \label{Kitchen-RL-table-2d}
  \centering
  \begin{tabular}{lrrrr}
 & \multicolumn{2}{c}{Baselines} & \multicolumn{2}{c}{MaskedDP} \\
 \cmidrule(lr){2-3} \cmidrule(lr){4-5}
    & 
    SingleRM-IQL    & VPL-OfflineRL-IQL   & Traj-MaskedDP-IQL  & CLS-MaskedDP-IQL \\
    \midrule
User 1  (Microwave) & 0.\second{548 ± 0.498}  &   0.326 ± 0.469 & \best{0.680 ± 0.466}   &  0.516 ± 0.500\\
User 2 (Kettle) & 0.00 ± 0.000 &  \second{0.400 ± 0.490}  & \best{0.988 ± 0.109} &  0.152 ± 0.359 \\
 \midrule
Average &0.274 ± 0.274&  \second{ 0.363 ± 0.037} & \best{0.834 ± 0.154
} &0.334 ± 0.182\\
    \bottomrule
  \end{tabular} 
  
\end{table*}

\begin{table*}[t!]
\centering
\footnotesize
\setlength{\tabcolsep}{1pt}
  \caption{\textbf{RL Performance on MO-Kitchen-3d} when evaluated over 100 episodes across 5 seeds reported in terms of mean +/- std for Task Success Rate. Since Franka-Kitchen is an offline environment, we run only offline RL IQL.}
  \label{Kitchen-RL-table-3d}
  \centering
  \begin{tabular}{lrrrr}
 & \multicolumn{2}{c}{Baselines} & \multicolumn{2}{c}{MaskedDP} \\
 \cmidrule(lr){2-3} \cmidrule(lr){4-5}
    & 
    SingleRM-IQL    & VPL-OfflineRL-IQL   & Traj-MaskedDP-IQL  & CLS-MaskedDP-IQL \\
    \midrule
User 1  (Microwave) & 0.342 ± 0.474  &  0.400 ± 0.490 & \second{0.528 ± 0.499}   &  \best{0.702 ± 0.457}\\
User 2 (Kettle) & 0.000 ± 0.000 &  0.002 ± 0.045  & \best{0.900 ± 0.300} &   \second{0.028 ± 0.165} \\
 User 3 (Light Switch) & \second{0.200 ± 0.400} & 0.000 ± 0.000  & \best{0.970 ± 0.171} &  0.00 ± 0.000 \\
 \midrule
Average &0.181 ± 0.140 &  0.134 ± 0.188 & \best{0.799 ± 0.194} & \second{0.243 ± 0.325} \\
    \bottomrule
  \end{tabular} 
  
\end{table*}

\begin{table*}[t!]
\centering
\footnotesize
\setlength{\tabcolsep}{1pt}
  \caption{\textbf{RL Performance on MO-Kitchen-4d} when evaluated over 100 episodes across 5 seeds reported in terms of mean +/- std for Task Success Rate. Since Franka-Kitchen is an offline environment, we run only offline RL IQL.}
  \label{Kitchen-RL-table-4d}
  \centering
  \begin{tabular}{lrrrr}
 & \multicolumn{2}{c}{Baselines} & \multicolumn{2}{c}{MaskedDP} \\
 \cmidrule(lr){2-3} \cmidrule(lr){4-5}
    & 
    SingleRM-IQL    & VPL-OfflineRL-IQL   & Traj-MaskedDP-IQL  & CLS-MaskedDP-IQL \\
    \midrule
User 1  (Microwave) & 0.000 ± 0.000  &\second{0.414 ± 0.493} & \best{0.526 ± 0.499}   &  0.178 ± 0.383 \\
User 2 (Kettle) & 0.000 ± 0.000 & 0.000 ± 0.000  & \best{ 0.738 ± 0.440} & \second{0.004 ± 0.063}\\
User 3 (Light Switch) & 0.000 ± 0.000 & 0.000 ± 0.000  & \best{ 0.998 ± 0.045}& \second{0.002 ± 0.045} \\
User 4  (Top Burner) & 0.000 ± 0.000 & 0.000 ± 0.000  & \best{0.384 ± 0.486} & 0.000 ± 0.000\\
 \midrule
Average &0.000 ± 0.000 &  \second{0.103 ± 0.179}  & \best{ 0.661 ± 0.232} & 0.046 ± 0.076\\
    \bottomrule
  \end{tabular} 
  
\end{table*}

%  \section{Additional experiments} \label{app:additional_exps_more_envs}
%  % This is where to add experiments on training time , num pca adaption queries , 
% We additionally report training time as mean and standard deviation across 5 seeds on NVIDIA GeForce RTX 3090  in Table~\ref{all-env-training-time-table} and we demonstrate that P3L substantially reduces compute cost and training time . 

% \begin{table*}
% \centering
% \footnotesize
%   \caption{\textbf{Training time in seconds when evaluated  reported as mean and standard deviation across 5 seeds}: the lower time it takes,
% the faster the framework is. P3L significantly reduces training time in all environments.}\label{all-env-training-time-table}
%   \centering
%   \begin{tabular}{lrrrr}
%  & \multicolumn{1}{c}{Baseline} & \multicolumn{2}{c}{P3L} \\
% \cmidrule(lr){2-2} \cmidrule(lr){3-4}
% Environment  & VPL & Privileged-Traj-MaskedDP & Anonymized-CLS-MaskedDP \\
% \midrule
% MO-Highway   &131.63 $\pm$ 3.19 &\best{0.75 $\pm$  0.49} &  \second{2.16 $\pm$ 0.20}   \\
% \midrule
%  MO-Reacher & 225.21 $\pm$ 9.33
% & \second{7.83 $\pm$ 0.97}
% & \best{4.16 $\pm$ 0.06} \\
% \midrule 
% MO-Humanoid  & 18018.25 $\pm$ 3784.13& \best{7.23 $\pm$ 0.88}   & \second{14.07 $\pm$ 3.55}\\
% % \midrule
% % Franka-Kitchen -DECIDE WRT SEEDS & &    & \\

% \bottomrule
% \end{tabular}

% \end{table*}
 
\section{OOD evaluation}\label{app:more_ood_eval}
% OOD for all other environments and combined objectives

\begin{figure*}[h]
     \centering
         \includegraphics[width=0.95\linewidth]{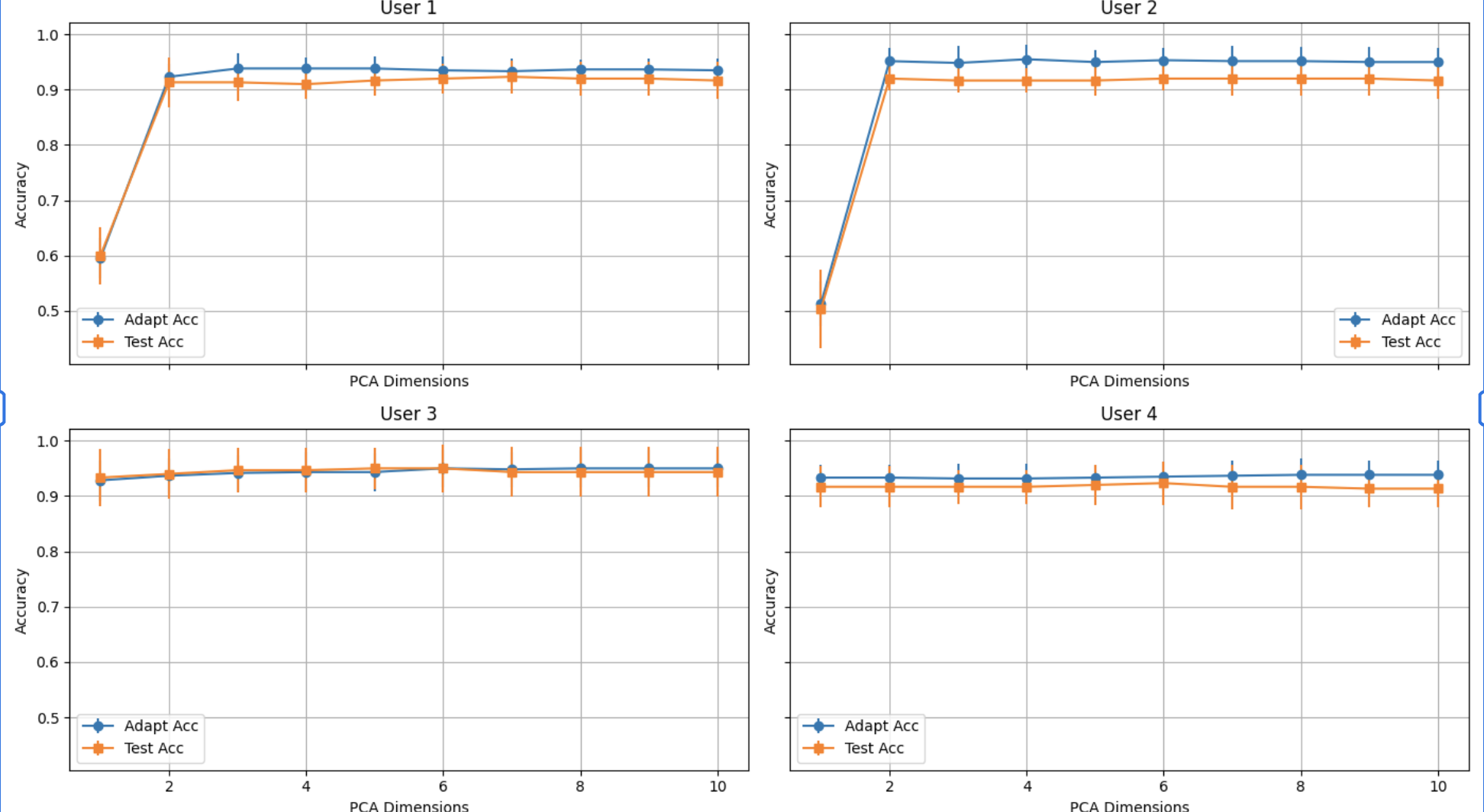}
         \caption{\textbf{Per User Test Accuracy vs Number of PCA basis eigenvectors} calculated per user across 10 seeds for each PCA size with standard deviation plotted as error bars. }
        \label{fig:pca_per_user}
\end{figure*}

\subsection{MO-Reacher}
We explore personalization to out-of-distribution (OOD) new user at test time by training on any possible set of 3 users in MO-Reacher and testing adaptation to a new user with an unseen behavior, both with respect to test set accuracy as well as the RL performance. 
For adaptation to a new user, to get $k_j$ for our frameworks and to infer latent preference variable for VPL, we use 30 adaptation queries.
And since this is an online adaptation to new user out of training distribution, we only optimize an RL ARS policy under VPL unlike in-distribution VPL-IQL.
We report mean and standard deviation across 5 seeds for 15 test preference queries and 100 rollouts for RL for the two metrics respectively in Table ~\ref{tab:OOD_Reacher}.
Our approach, Traj-MaskedDP outperforms VPL on both test set accuracy as well as on downstream RL performance for 4 different OOD users as well as on average, while making same assumptions about access to annotator identity in offline diverse dataset. 

\begin{table*}[h]
% \footnotesize
\scriptsize
\setlength{\tabcolsep}{1pt}
% \caption{\textbf{Out of Distribution Adaptation to New Users} reported as Mean and standard deviation calculated over 2 seeds for 15 test preference queries and 100 rollouts for Test time New User Accuracy and downstream RL performance respectively.}
\caption{\textbf{Out-of-distribution adaptation to new users on MO-Reacher}. Results report mean ± std over 5 seeds for test preference accuracy (15 queries) and RL performance (100 evaluation rollouts). Our methods generalize better to unseen users, achieving higher preference accuracy and stronger downstream RL performance than VPL.}
\label{tab:OOD_Reacher}
\centering
\begin{tabular}{lllrrr}
\toprule
\multicolumn{3}{c}{Setup} & \multicolumn{1}{c}{Baselines} & \multicolumn{2}{c}{MaskedDP} \\
\cmidrule(lr){1-3} \cmidrule(lr){4-4} \cmidrule(lr){5-6}
  OOD New User& Trained on Users&Eval Metric  & VPL & Traj-MaskedDP & CLS-MaskedDP \\
\midrule
% \multirow{2}{*}{1} & \multirow{2}{*}{2-3-4} &  Test &   65.333 ±2.667    & \best{100 ± 0.00 }        & \best{100 ± 0.00 }      \\
%                       &   & RL   & &  &  \\  
\multirow{2}{*}{1} & \multirow{2}{*}{2-3-4} &  Test &  36.000±3.266 & \best{100 ± 0.00 }        & \best{100 ± 0.00 }      \\
                      &   & RL   &  22.048 ± 1.975 & \best{32.841 ± 3.033} & \second{29.787 ± 3.328} \\  
                      \midrule
\multirow{2}{*}{2} & \multirow{2}{*}{1-3-4} &  Test &   33.333±0.000   & \best{100 ±0.00 }        &  \best{100 ±0.00 }    \\
      &   & RL   & 17.255 ± 2.414 & \best{31.249 ±  1.178}  & \second{21.401 ± 4.460}  \\
      \midrule
\multirow{2}{*}{3} & \multirow{2}{*}{1-2-4} &  Test &  0.000±0.000     & \best{100.00±0.00}       &   77.333±3.266 \\
                      &   & RL   & 21.637 ± 2.498 & \best{31.089 ± 3.943}  & \second{24.611 ±  10.784} \\
                      \midrule

\multirow{2}{*}{4} & \multirow{2}{*}{1-2-3} &  Test & \best{100.00±0.00}      &\best{100.00±0.00}    &  \best{100.00±0.00}   \\
                      &   & RL   & 17.450 ± 2.820 & \best{35.028 ± 3.195 } &  \second{15.069 ± 5.099} \\
 \midrule                     

\multirow{2}{*}{Average} & \multirow{2}{*} & Test &    42.333 ± 36.189 & \best{100.00±0.00}    &   \second{94.333 ± 9.815}  \\
                      &   & RL   &  19.598 ± 2.251 &  \best{32.552 ± 1.585} & \second{22.717 ± 5.334} \\                           
\bottomrule
\end{tabular}

\end{table*}

\subsection{MO-Highway}
Similar to MO-Reacher above, we explore OOD to MO-Highway using 30 adaptation preference queries and PCA dim of 3 as used before during the In-Distribution experiments and report RL performance over 100 rollouts in Table ~\ref{tab:OOD_Highway}.
We observe that P3L is able to provide significantly higher personalization and alignment with respect to both metrics at test time, even when generalizing to out-of-distribution new users.

\begin{table*}[h]
% \footnotesize
\scriptsize
\setlength{\tabcolsep}{1pt}
\caption{\textbf{Out-of-distribution adaptation to new users on MO-Highway}. Results report mean ± std over 5 seeds for test preference accuracy (10 queries) and RL performance (100 evaluation rollouts). Our methods generalize better to unseen users, achieving higher preference accuracy and stronger downstream RL performance than VPL.}

\label{tab:OOD_Highway}
\centering
\begin{tabular}{lllrrr}
\toprule
\multicolumn{3}{c}{Setup} & \multicolumn{1}{c}{Baselines} & \multicolumn{2}{c}{MaskedDP} \\
\cmidrule(lr){1-3} \cmidrule(lr){4-4} \cmidrule(lr){5-6}
  OOD New User& Trained on Users&Eval Metric  & VPL & Traj-MaskedDP & CLS-MaskedDP \\
\midrule
% \multirow{2}{*}{1} & \multirow{2}{*}{2-3-4} &  Test &   65.333 ±2.667    & \best{100 ± 0.00 }        & \best{100 ± 0.00 }      \\
%                       &   & RL   & &  &  \\  
\multirow{2}{*}{1} & \multirow{2}{*}{2-3} &  Test & 0.000±0.000 & \second{ 68.000±7.483 }        & \best{74.000±21.541 }      \\
                      &   & RL   &  \best{5.603 ± 3.621 }& \second{5.389 ± 3.788} &  5.109 ± 3.490 \\  
                      \midrule

\multirow{2}{*}{2} & \multirow{2}{*}{1-3} &  Test &  68.000±9.798   & \best{92.000±7.483 }        &  \second{90.000±10.954}    \\
      &   & RL   & 0.680 ± 2.125 & \best{29.836 ± 0.235 }  & \second{ 29.833 ± 0.242}  \\
      \midrule

\multirow{2}{*}{3} & \multirow{2}{*}{1-2} &  Test & 60.000±16.733    & \best{84.000±10.198}       &   \second{ 70.000±23.664} \\
                      &   & RL   & -22.092 ± 6.159 & \second{-20.628 ± 6.971} &\best{-13.586 ± 10.540} \\
                      \midrule

\multirow{2}{*}{Average} & \multirow{2}{*} & Test &  42.667 ± 30.346 
& \best{81.333 ± 9.978}
& \second{78.000 ± 8.641} \\

                      & & RL & -5.270 ± 12.064 & \second{4.866 ± 20.605}  & \best{ 7.119 ± 17.783}\\                           
\bottomrule
\end{tabular}
\end{table*}

\section{Ablation on PCA dimensions}\label{app:pca_dim_ablation}
\begin{figure*}
     \centering
         \includegraphics[width=0.75\linewidth]{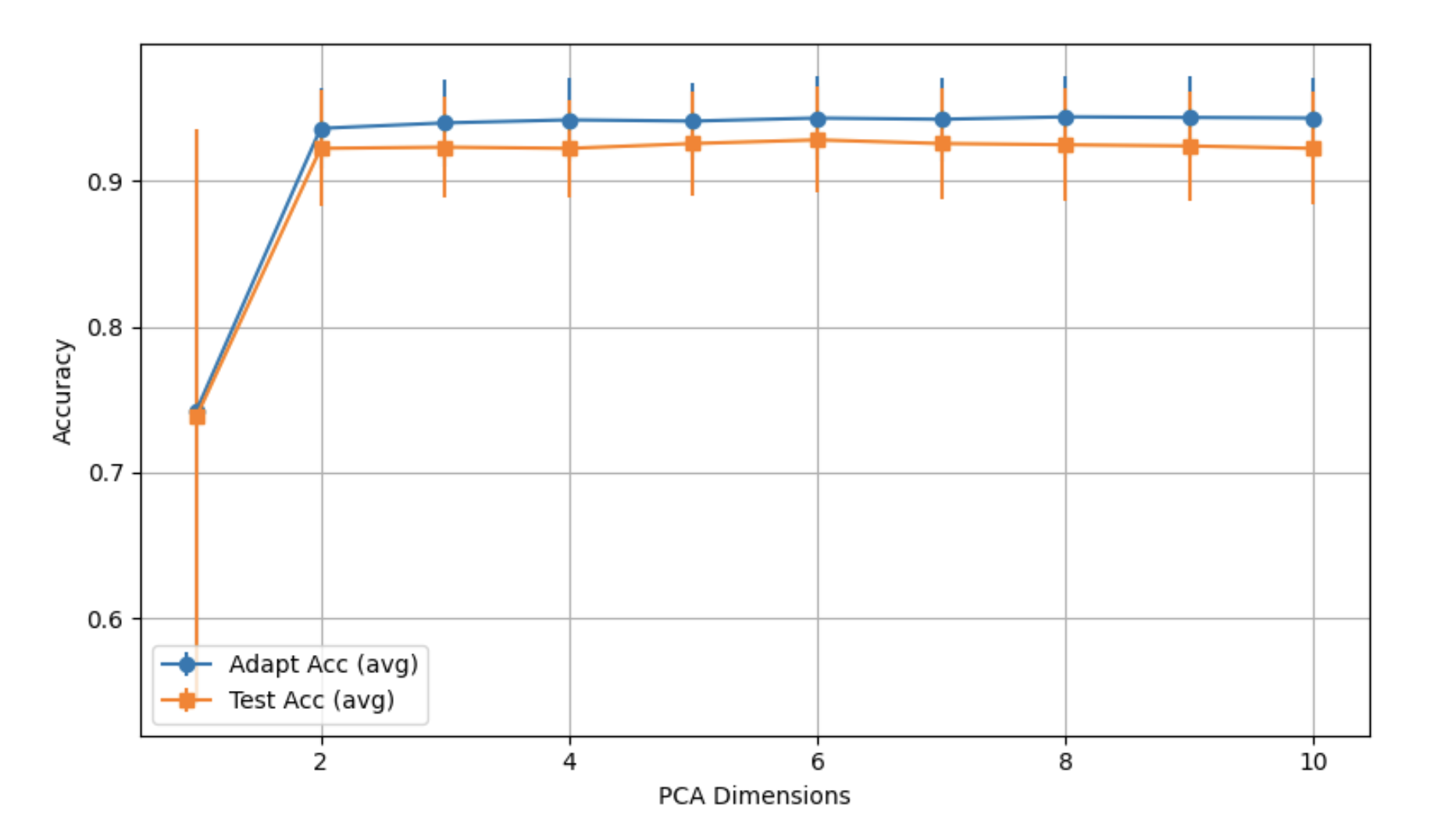}
         \caption{\textbf{Average Test Accuracy vs Number of PCA basis eigenvectors} calculated across all users over 10 seeds for each PCA size  with standard deviation plotted as error bars.}
        \label{fig:pca_avg_user}
\end{figure*}

We show test-time accuracy as the number of PCA dimensions is varied for each individual user in Fig. ~\ref{fig:pca_per_user} and averaged across all users in Fig. ~\ref{fig:pca_avg_user}, respectively. For all PCA dimensions, Traj-MaskedDP PCA is trained with 600 queries, adapted with 60 and tested with a fixed set of 30 queries per user using Traj-MaskedDP.

\section{Ablation Per-User Adaptive queries}\label{app:per_user_adapative_ablation}
All models are trained with 600 preference queries and tested on fixed set of 30 queries per user.
We observe similar trend as we do for average test time accuracy vs adaptation query size, that our frameworks can do comparably high even with smaller adaptation number of queries as they can with 100$x$ queries for adaptation whereas VPL is very sensitive to number of queries used for inferring its latent preference . 
\begin{figure*}
     \centering
          \includegraphics[width=0.95\linewidth]{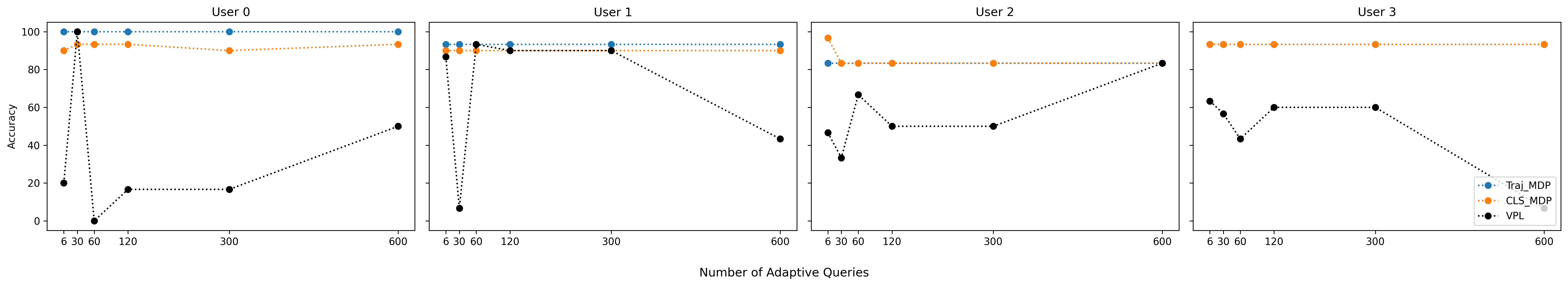}

         \caption{\textbf{Per-User Test Time Accuracy vs Number of adaptation queries} where each user favors a distinct goal in the environment.}
        \label{fig:pipeline}
\end{figure*}

\section{Ablation on Training queries}\label{app:num_training_queries_ablation}
\begin{figure*}
     \centering
         \includegraphics[width=0.45\linewidth]{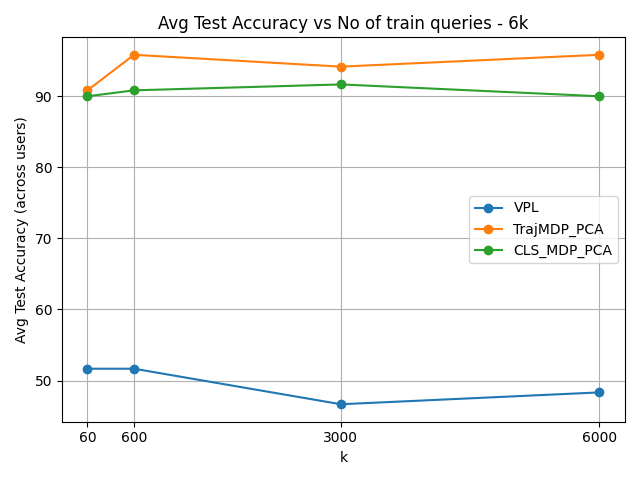}
         \caption{\textbf{Training Query Size Ablation}}
        \label{fig:training query ablation}
\end{figure*}

All models are adapted with 60 preference queries and tested on fixed set of 30 queries per user. Our framework in both settings outperforms VPL as shown in Fig ~\ref{fig:training query ablation}

\section{VPL Posterior Collapse}\label{app:vpl_posterior_collapse}

We observed the encoder in VPL, Variational Autoencoder (VAE), begins to produce NaN values for the latent preference mean $z$ in two environments: MO-Humanoid and Franka-Kitchen.
For both environments we report parameters that originally resulted in NaN values of the latent mean and thereby collapsing training of the VPL reward model (which is actually the decoder with input as observation and the latent mean produced by the VAE).

We performed extensive hyperparameter tuning and found that reducing annotation set size, also termed as context set size in ~\cite{vpl_poddar2024personalizing}, often led to more stable training.
% For Kitchen, when training VPL on heterogeneous preference dataset containing trajectories from all tasks with  training pairwise preference 2400 queries and since each query can be annotated by any possible user preferring any task this leads to $4x2400$ labels, we observed this for seed 0, annotation set size 30,trajectory length of 280, and adaptation preference dataset of size 180. 
For Kitchen, the VPL training dataset is heterogeneous, combining trajectories across all tasks and user preferences.
The training dataset contains 2,400 pairwise preference queries, and as each query can be annotated from the perspective of a user who prefers any one of the possible tasks, each query yields one label per possible preferred task, leading to 9600 preference labels.
 We observed this collapse for seed 0,  annotation set size 30 where each trajectory in pairwise preference query is of length 280, and an adaptation preference dataset of size 180.
For MO-Humanoid, we observed this when sweeping over annotation set size of 25 as well as with 5, over 1000 preference queries of length 150.

% \bibliography{aaai2027}

% Check whether the conference requires a reproducibility checklist to be included in the paper.
% If so, you can uncomment the following line and ajust the path to include it.
% \input{ReproducibilityChecklist.tex}

\end{document}